%% file: paper.tex
\documentclass[sigconf]{acmart}
\renewcommand\footnotetextcopyrightpermission[1]{} 

\usepackage{booktabs} 
\usepackage{url}
\usepackage{amsmath}
\usepackage{amssymb}
\usepackage{graphicx}
\usepackage{hyperref}
\usepackage{wrapfig}

\usepackage{cleveref}
\usepackage[bottom]{footmisc}

\usepackage{listings}
\usepackage{color}

\definecolor{dkgreen}{rgb}{0,0.6,0}
\definecolor{gray}{rgb}{0.5,0.5,0.5}
\definecolor{mauve}{rgb}{0.58,0,0.82}

\setcopyright{none}

\acmConference[]{}{}{}
\acmYear{2017}
\copyrightyear{2017}

\begin{document}
\title{Statistical Analysis on E-Commerce Reviews, with Sentiment Classification using Bidirectional Recurrent Neural Network}

\author{Abien Fred M. Agarap}
\email{abienfred.agarap@gmail.com}

\begin{abstract}
Understanding customer sentiments is of paramount importance in marketing strategies today. Not only will it give companies an insight as to how customers perceive their products and/or services, but it will also give them an idea on how to improve their offers. This paper attempts to understand the correlation of different variables in customer reviews on a women clothing e-commerce, and to classify each review whether it recommends the reviewed product or not and whether it consists of positive, negative, or neutral sentiment. To achieve these goals, we employed univariate and multivariate analyses on dataset features except for review titles and review texts, and we implemented a bidirectional recurrent neural network (RNN) with long-short term memory unit (LSTM) for recommendation and sentiment classification. Results have shown that a recommendation is a strong indicator of a positive sentiment score, and vice-versa. On the other hand, ratings in product reviews are fuzzy indicators of sentiment scores. We also found out that the bidirectional LSTM was able to reach an F1-score of 0.88 for recommendation classification, and 0.93 for sentiment classification.
\end{abstract}



\maketitle

\input{body}

\end{document}

%% file: body.tex
\section{Introduction}
Companies are starting to turn to social media listening as a tool for understanding their customers, in order to further improve their products and/or services. As a part of this movement, text analysis has become an active field of research in computational linguistics and natural language processing.\\
\indent	One of the most popular problems in the mentioned field is text classification, a task which attempts to categorize documents to one or more classes that may be done manually or computationally. Towards this direction, recent years have shown top interest in classifying \textit{sentiments} of statements found in social media, review sites, and discussion groups. This task is known as \textit{sentiment analysis}, a computational process that uses statistics and natural language processing techniques to identify and categorize opinions expressed in a text, particularly, to determine the polarity of attitude (positive, negative, or neutral) of the writer towards a topic or a product\cite{emc2015data}. The said task is now widely used by companies for understanding their clients through their customer support in social media, or through their review boards online.\\
\indent	In this paper, we attempt to analyze the customer reviews on women clothing e-commerce\cite{brooks2018women} by employing statistical analysis and sentiment classification. We first analyze the non-text review features (e.g. age, class of dress purchased, etc.) found in the dataset, as an attempt to unravel any connection between them and customer recommendation on the product. Then, we implement a bidirectional recurrent neural network (RNN) with long-short term memory (LSTM)\cite{hochreiter1997long} for classifying whether a review text recommends the purchased product or not, and for classifying the user review sentiment towards the product.

\section{Methodology}

\subsection{Machine Intelligence Library}
Keras\cite{chollet2015keras} with Google TensorFlow\cite{tensorflow2015-whitepaper} was used to implement the bidirectional recurrent neural network (RNN) with long-short term memory (LSTM)\cite{hochreiter1997long} in this study. As for the data preprocessing and handling, the \texttt{numpy}\cite{walt2011numpy} and \texttt{pandas}\cite{mckinney-proc-scipy-2010} Python libraries were used. Lastly, for the data visualization, the \texttt{matplotlib}\cite{Hunter:2007} and \texttt{seaborn}\cite{michael_waskom_2017_883859} Python libraries were used.

\subsection{The Dataset}
The Women's Clothing E-Commerce Reviews\cite{brooks2018women} was used as the dataset for this study. This dataset consists of reviews written by real customers, hence it has been anonymized, i.e. customer names were not included, and references to the company were replaced with ``retailer'' by \cite{brooks2018women}.\\
\indent	Table \ref{table:freq-dist} shows the frequency distribution of dataset features and label (\texttt{Recommended IND}).

\begin{table}[htb!]
\centering
\caption{Frequency Distribution of Dataset Features.}
		\begin{tabular}{cc}
		\toprule
		Feature		&	Unique Count\\
		\midrule
		Clothing ID & 1172\\
		Age & 77\\
		Title  & 13984\\
		Review Text & 22621\\
		Rating & 5\\
		Recommended IND & 2\\
		Positive Feedback Count & 82\\
		Division Name & 3\\
		Department Name & 6\\
		Class Name & 20\\
		\bottomrule
		\end{tabular}\\
		\label{table:freq-dist}
\end{table}

\subsection{Data Analysis}

In which we analyze the dataset features and their implications on user recommendation and review sentiments. This subsection covers four statistical analyses. Table \ref{table:stat-desc} shows a summary of statistical description of the dataset.

\begin{table}[htb!]
\centering
\caption{Summary of Statistical Description of Dataset Features.}
	\begin{tabular}{cccc}
	\toprule
	Feature & Mean & Standard Deviation & Type\\
	\midrule
	Clothing ID & 919.695908 & 201.683804 & Integer\\
	Age & 43.282880 & 12.328176 & Integer\\
	Rating & 4.183092 & 1.115911 & Categorical\\
	Recommended & 0.818764 & 0.385222 & Categorical\\
	Positive Feedback & 2.631784 & 5.787520 & Integer\\
	\bottomrule
	\end{tabular}\\
	\label{table:stat-desc}
\end{table}

\subsubsection{Analysis on Univariate Distributions}

\begin{enumerate}
	\item \textbf{Age and Positive Feedback Count.} Figure \ref{age-distribution} reveals that the most engaged customers in reviewing purchased products were in the age range of 35 to 44. In addition, the figure suggests that they have the most positive reviews on their purchased products. From this, we have two points to consider: (1) the said age group is the most satisfied group in the range of customers, thus, the e-commerce at review must focus on maintaining this segment, and (2) the e-commerce entity can explore why other age groups are comparatively less satisfied than the age group 35 to 44.
	\item \textbf{Department Name and Division Name.} Figure \ref{divname-and-deptname-freqdist} shows the frequency distribution of customer reviews per \textit{department} and per \textit{division}. This gives the e-commerce an insight on the customer apparel sizes and clothing being most reviewed, i.e. \textit{General} which refers to clothing size, and \textit{tops} which refers to apparel types.
	\item \textbf{Top 60 Clothing ID.} Figure \ref{top-60} shows the IDs of top 60 reviewed apparel from the e-commerce. The apparels with clothing IDs 1078, 862, and 1094 belong to the \textit{general} division and \textit{dresses} apparely type, with a positive title review of ``Beautiful dress'' as per \cite{brooks2018guided}.
	\item \textbf{Class Name.} Figure \ref{classname-freqdist} shows the frequency distribution of apparel classes most reviewed. The top three apparels are \textit{dresses}, \textit{knits}, and \textit{blouses}.
	\item \textbf{Rating, Recommendation, and Label.} Figure \ref{rating-recommended-label} shows that the dominant reviews were positive, suggesting that the e-commerce fairly satisfies its customers. It may be axiomatic that a review with recommendation implies a higher rating and a positive sentiment. But then again, the processing of sentiments were based on a threshold of higher than rating of 3 for positive, and negative for the rest. We shall look more into this in subsection \ref{sentiment-analysis}.
	\item \textbf{Word Length.} Figure \ref{word-length} shows that regardless of the rating in a review, apparel type, or recommendation, the users had qualitatively the same length of words in their reviews.
\end{enumerate}

\subsubsection{Analysis on Multivariate Distributions}

\begin{enumerate}
	\item \textbf{Division Name by Department Name.} Figure \ref{divname-deptname} reveals the dominance of general-sized tops, while Figure \ref{divname-deptname-pivot} supports this inference.
	\item \textbf{Class Name by Department Name.} Figure \ref{classname-deptname} reveals the dominance of dress among apparel types, and supported by Figure \ref{classname-deptname-pivot}.
	\item \textbf{Class Name by Division Name.} Figure \ref{classname-divname} reveals the most reviewed apparel types as general-sized blouses, dresses, and knits. However, \ref{classname-divname-pivot} shows that most reviews on dresses are from general petite sizes.
	\item \textbf{Age by Positive Feedback Count.} Figure \ref{age-positivefeedback-scatter} shows a small correlation between age and the positive feedback in a review. Based on the figure, the same age group of 35 to 44 seems to be the group that gave most of the positive feedbacks.
	\item \textbf{Recommendation by Department Name and Division Name.} Figure \ref{recommendation-deptname-divname} corroborates the findings in Figure \ref{divname-deptname}.
	\item \textbf{Rating by Department Name and Division Name.} Figure \ref{rating-deptname-divname} shows consistency in rating distribution.
	\item \textbf{Rating by Recommendation.} Figure \ref{rating-recommended} supports the assumption that a review rating mirrors its recommendation status, i.e. higher rating means recommendation and vice-versa.
\end{enumerate}

\subsubsection{Multivariate Analysis and Descriptive Statistics}

\begin{enumerate}
	\item \textbf{Average Rating by Recommendation.} Figure \ref{averagerating-deptname-recommended} shows consistency on recommendation and rating, i.e. when review has recommendation, the rating is under maximum value of rating; when review has no recommendation, the rating is halved.
	\item \textbf{Average Rating and Recommendation by Clothing ID Correlation.} Figure \ref{meanrating-recommended-clothing-corr} attempts to look at the correlation, if there is any, between the average rating of a product and number of reviews for a product, that is grouped by clothing ID. The correlation matrix suggests there is no such correlation between the variables considered, but it did reveal a relatively strong correlation of 0.8 between rating and recommendation. The mentioned correlation coefficient further substantiates the assumption on connection between rating and recommendation.
\end{enumerate}

\subsubsection{Word Frequency Distributions}

\begin{enumerate}
	\item \textbf{Titles.} Figure \ref{wordcloud-titles} gives us the most frequent words in a review title. Only the word ``flaws'' seems to indicate a negative review, but then again, this does not necessarily indicate that the entire product review has a negative sentiment. Take not that this word cloud only accounts for the frequency of words in titles, and does not account for phrases. In other words, there may be counter-words for negative word indicators, but only failed to make it into the word cloud. The same may be said for positive word indicators in the word cloud, as it does not include any negators if there are any.
	\item \textbf{Most Frequent Words in Highly-rated Comments.} Since Figure \ref{wc-most-freq-words-high-rate-comments} is a word cloud for reviews with high ratings, it may be assumed that the words in this figure reflects what are written in their respective reviews.
	\item \textbf{Most Frequent Words in Low-rated Comments.} Since Figure \ref{most-freq-words-low-rate-comment} is a word cloud for reviews with low ratings, it may be assumed that the words in this figure reflects what are written in their respective reviews.
	\item \textbf{Word Clouds for Division Names.} Figure \ref{wordcloud-intimates} shows the most frequent words in product reviews from the ``intimates'' division; Figure \ref{wordcloud-general} for ``general'' division; and Figure \ref{wordcloud-petite} for ``general petite'' division. Further investigation on these word clouds may reveal some useful insight on customer acceptability per division. 
\end{enumerate}

\subsection{Dataset Preprocessing}

\subsubsection{Text Cleaning}

The user review texts were cleaned by eliminating delimiters such as \texttt{\textbackslash{n}} and \texttt{\textbackslash{r}} found in the texts.

\subsubsection{Sentiment Analysis}\label{sentiment-analysis}

Instead of manually tagging the review texts, the sentiment analyzer of NLTK\cite{Loper02nltk:the} was used to automate the process. Thus, leaving behind the intuitive tagging of review texts that had a threshold of rating 3, i.e. if a review rating is greater than or equal to 3, it is considered as a positive feedback, otherwise it is considered a negative feedback. The mentioned manual, intuitive tagging had the flaw of not taking into account some neutral sentiments. Hence, the use of sentiment analyzer by NLTK\cite{Loper02nltk:the}. See Figure \ref{norm-sentimentdist} for the frequency distribution of sentiments per recommendation.

\subsubsection{Word Embeddings}

The GloVe word embeddings\cite{pennington2014glove} were used to map the words in review texts to the vector space.

\subsection{Machine Learning}

\begin{figure}[!htb]
\minipage{0.5\textwidth}
\centering
	\includegraphics[width=\textwidth]{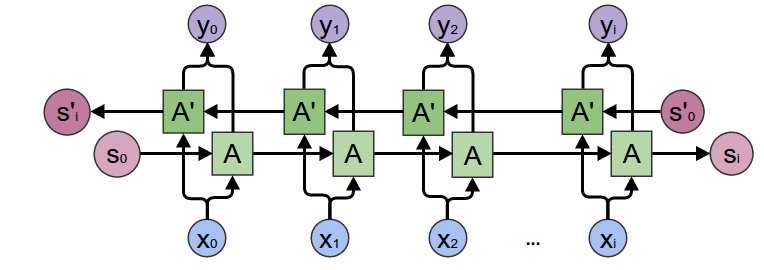}
	\caption{Image from \cite{olah2015neural}. Computation of a conventional Bidirectional RNN maps input sequences $x$ to target sequences $y$, with loss $L(t)$ at each time step $t$. The RNN cells $s$ propagate information forward in time (towards the right) while the RNN cells $s'$ propagate information backward in time (towards the left). Thus at each time step $t$, the output units $o(t)$ (before applying an activation function to get $y$) can benefit from a relevant summary of the past in its $s(t)$ input, and from a relevant summary of the future in its $s'(t)$ input.}
	\label{bidirectional-rnn-lstm}
\endminipage\hfill
\end{figure}

Given that the problem at hand is a classification task on sentiments, the most appropriate ML algorithm to implement is a recurrent neural network (RNN). However, from literature, we know that a vanilla RNN suffers from \textit{vanishing gradients}. Hence, we used the RNN with long-short term memory (LSTM) units, which was designed to solve the mentioned problem\cite{hochreiter1997long}. Furthermore, to better capture the context of words in the review texts, we employed a bidirectional RNN with LSTM (see Figure \ref{bidirectional-rnn-lstm}). That is, the model has the capability to learn the context from ``past'' to the ``future'' of a text sequence and vice-versa\cite{Goodfellow-et-al-2016}. In turn, giving the model more insight on each review text.\\
\indent Below are the LSTM gate equations\cite{hochreiter1997long}, which we implemented using Google TensorFlow\cite{tensorflow2015-whitepaper}.
\begin{align}
f_{t} = \sigma(W_{f} \cdot [h_{t - 1}, x_{t}] + b_{f})\\
i_{t} = \sigma(W_{i} \cdot [h_{t - 1}, x_{t}] + b_{i})\\
\tilde{C}_{t} = tanh(W_{C} \cdot [h_{t - 1}, x_{t}] + b_{C})\\
C_{t} = f_{t} * C_{t - 1} + i_{t} * \tilde{C}_{t}\\
o_{t} = \sigma(W_{o} \cdot [h_{t - 1}, x_{t}] + b_{o})\\
h_{t} = o_{t} * tanh(C_{t})
\end{align}
where $f$ is the \textit{forget gate}, which ``forgets'' non-essential information for the model; $i$ is the \textit{input gate}, which accepts new data input at a given time step $s_{t}$; $\tilde{C}$ is the candidate cell state value of each LSTM cell; $C$ is the cell state value to be passed onto the next RNN-LSTM cell; $o$ is the \textit{output gate} which decides what the cell state will output; and $h$ is the cell state output from cell state value and the decided output.\\
\indent We employed this machine learning model on two text classification problems on the dataset: (1) recommendation classification, which determines whether a review text recommends the reviewed product, and (2) sentiment classification, which determines the tone of the review text towards the purchased product.

\subsubsection{Recommendation Classification}
A product review has two recommendation states: (1) recommended, and (2) not recommended -- a binary classification problem.
\subsubsection{Sentiment Classification}
A product review has three sentiment states: (1) negative, (2) neutral, and (3) positive -- a multinomial classification problem.

\section{Results and Discussion}
All experiments in this study were conducted on a laptop computer with Intel Core(TM) i5-6300HQ CPU @ 2.30GHz x 4, 16GB of DDR3 RAM, and NVIDIA GeForce GTX 960M 4GB DDR5 GPU. The dataset was partitioned into a 60/20/20 fashion, i.e. 60\% for training dataset, 20\% for validation dataset, and 20\% for testing dataset.\\
\indent Table \ref{table:hyperparameters} shows the hyper-parameters used by the Bidirectional RNN-LSTM in the experiments, these hyper-parameters were arbitrarily chosen as hyper-parameter tuning implies more computational cost requirement. Table \ref{table:test-accuracy-loss} shows the test accuracy and test loss by the Bidirectional RNN-LSTM on both recommendation classification and sentiment classification experiments.

\begin{table}[!htb]
\centering
\caption{Hyper-parameters used in Bidirectional RNN-LSTM}
		\begin{tabular}{cc}
		\toprule
		Hyper-parameter		&	Value\\
		\midrule
		Batch Size & 256\\
		Cell Size & 256\\
		Dropout Rate & 0.50\\
		Epochs & 32\\
		Learning Rate & 1e-3\\
		\bottomrule
		\end{tabular}\\
		\label{table:hyperparameters}
\end{table}

\begin{table}[!htb]
\centering
\caption{Test Accuracy and Test Loss using Bidirectional LSTM.}
		\begin{tabular}{ccc}
		\toprule
		Task		&	Test Accuracy	&	Loss\\
		\midrule
		Recommendation Classification	&	$\approx$0.882678	&	$\approx$0.572342\\
		Sentiment Classification	&	$\approx$0.928414	&	$\approx$0.453205\\
		\bottomrule
		\end{tabular}\\
		\label{table:test-accuracy-loss}
\end{table}

However, take note that the frequency distributions for classes in \textit{recommendation} and \textit{sentiment} are both imbalanced, i.e. there are more \textit{recommended} classes than \textit{not recommended}, and there are more \textit{positive} sentiments than there are \textit{negative} and \textit{neutral} combined. This poses a problem as the model shall grow a biased classification towards the class with highest frequency distribution. Hence, we take a look at the statistical report on \textit{recommendation classification} at Table \ref{table:statistical-report-recommendation}.

\begin{table}[!htb]
\centering
\caption{Statistical Report on Recommendation Classification using Bidirectional LSTM.}
		\begin{tabular}{ccccc}
		\toprule
		Class		&	Precision	&	Recall	&	F1-Score	&	Support\\
		\midrule
		(0) Not Recommended & 0.70	&	0.65	&	0.68	&	847\\
		(1) Recommended & 0.92	&	0.94	&	0.93	&	3679\\
		Average / Total	&	0.88	&	0.88	&	0.88	&	4526\\
		\bottomrule
		\end{tabular}\\
		\label{table:statistical-report-recommendation}
\end{table}

\begin{figure}[!htb]
 \minipage[b]{0.4\textwidth}
 \centering
 	\includegraphics[width=\textwidth]{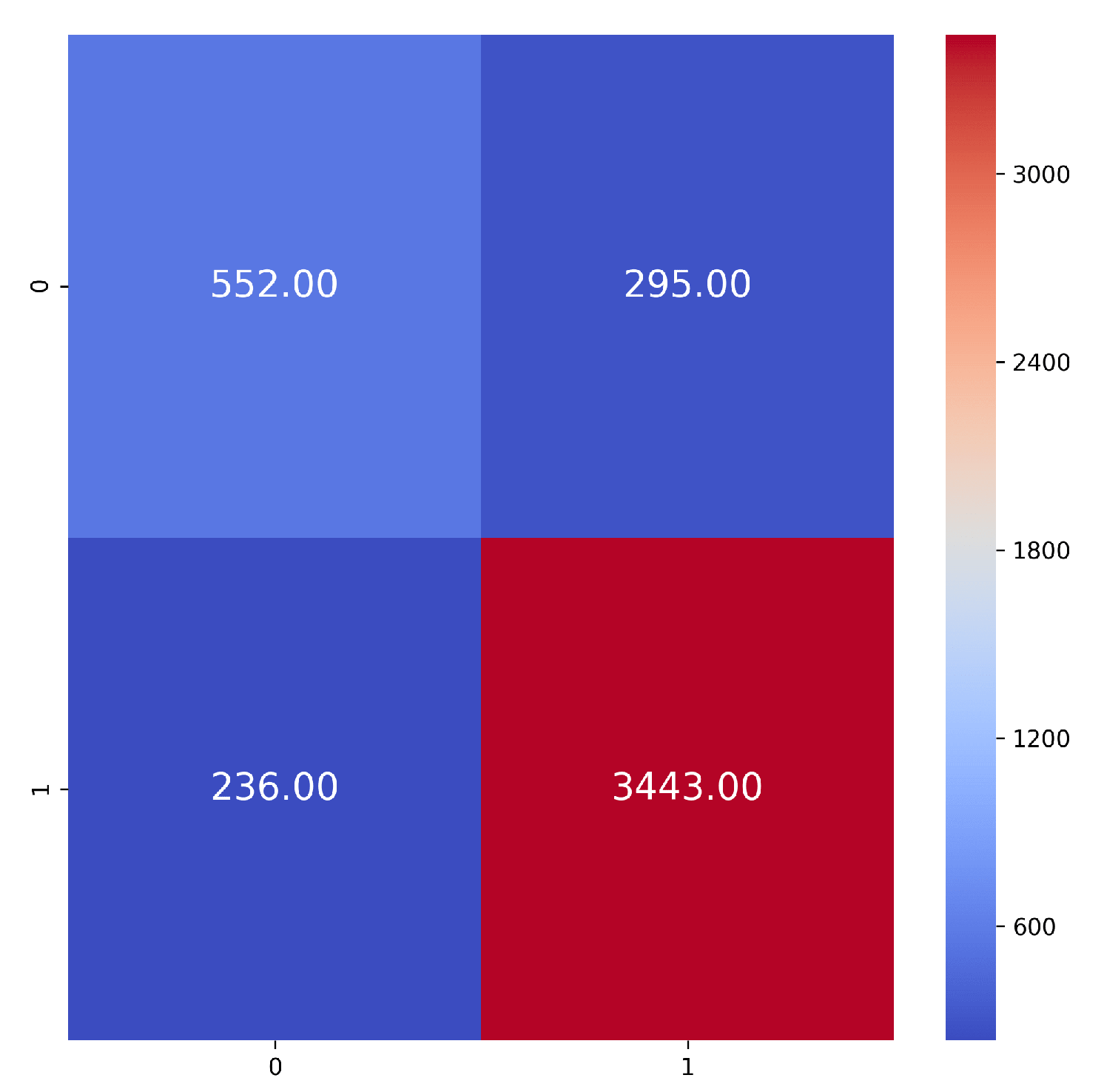}
 	\caption{The confusion matrix on recommendation classification.}
 	\label{conf_matrix_recommendation}
\endminipage\hfill
 \minipage[b]{0.4\textwidth}
 \centering
 	\includegraphics[width=\textwidth]{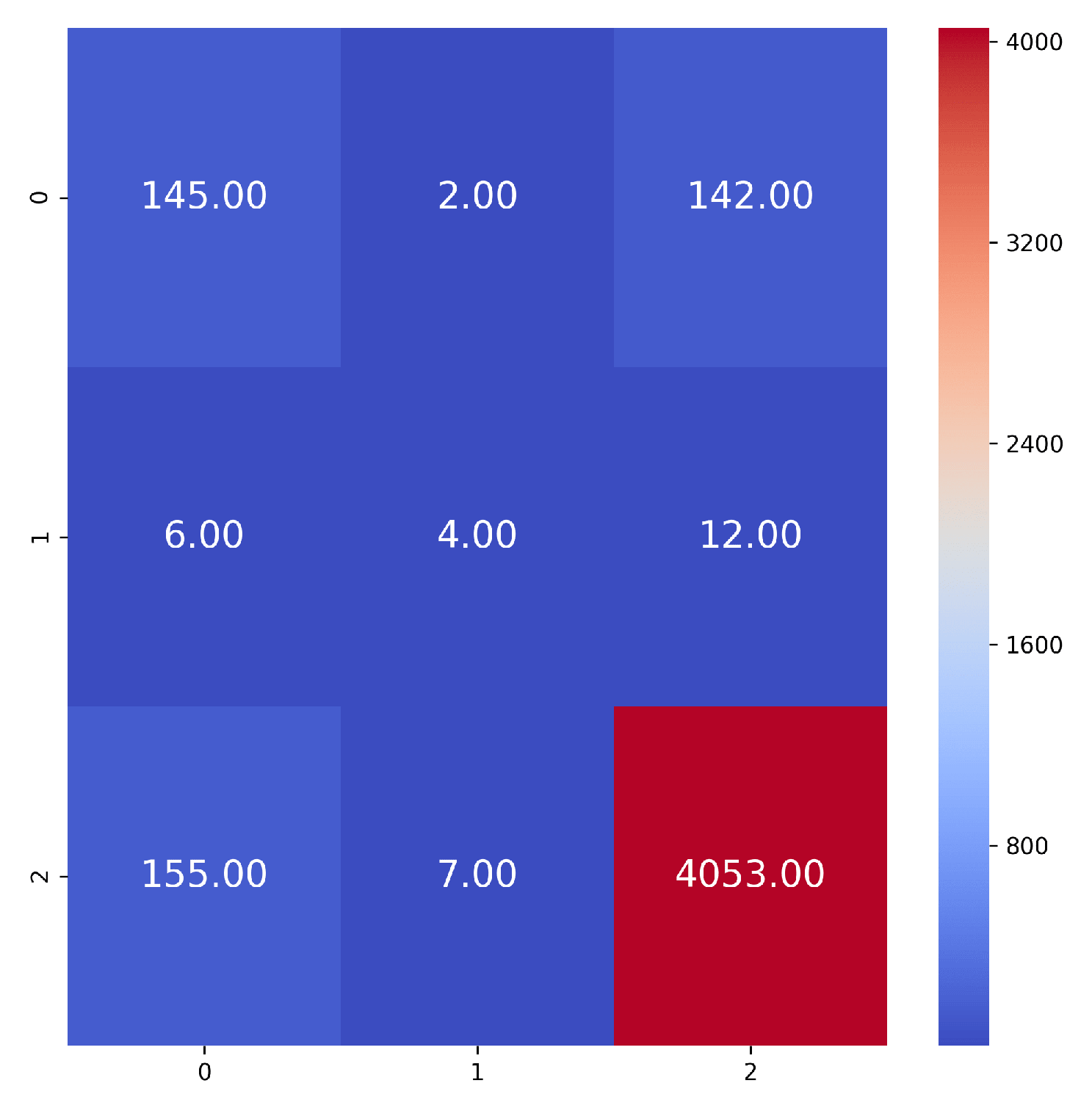}
 	\caption{The confusion matrix on sentiment classification.}
 	\label{conf_matrix_sentiment}
\endminipage\hfill
\end{figure}

\begin{figure}[!htb]
 \minipage{0.45\textwidth}
 \centering
 	\includegraphics[width=\textwidth]{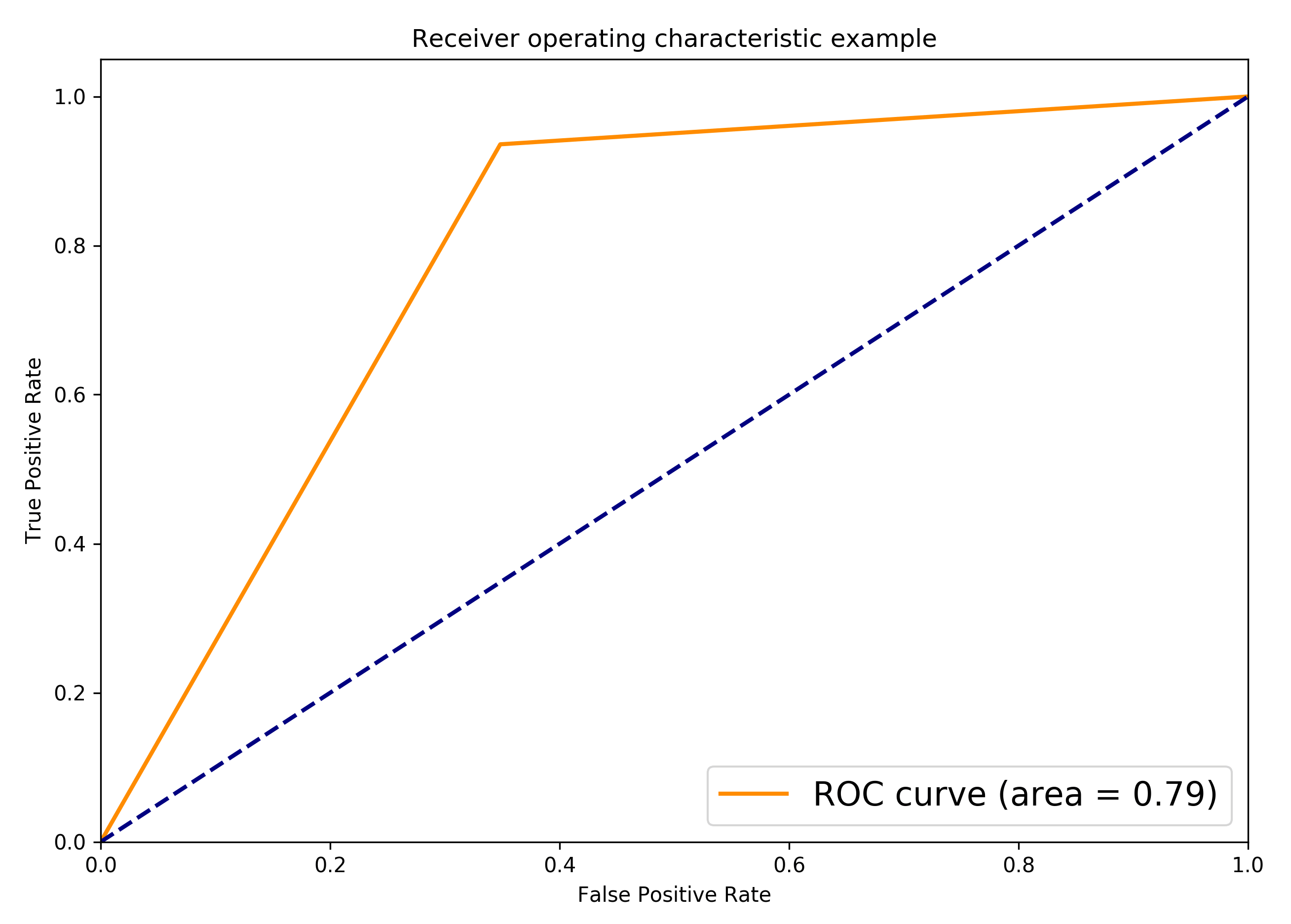}
 	\caption{The ROC Curve for binary classification on recommendation indicator.}
 	\label{roc-curve}
\endminipage\hfill
\end{figure}

Table \ref{table:statistical-report-recommendation} shows a relatively weaker predictive performance for negative class in the \textit{recommendation classification} problem, as it can also be seen in the confusion matrix in Figure \ref{conf_matrix_recommendation} (where $0$ represents \textit{not recommended} class, and $1$ represents \textit{recommended} class), thus supporting our claim above. To look at the model performance on a relatively fair scheme, we take a look at the ROC curve for the result (see Figure \ref{roc-curve}).

\begin{table}[!htb]
\centering
\caption{Statistical Report on Sentiment Classification using Bidirectional LSTM.}
		\begin{tabular}{ccccc}
		\toprule
		Class		&	Precision	&	Recall	&	F1-Score	&	Support\\
		\midrule
		(0) Negative & 0.47	&	0.50	&	0.49	&	289\\
		(1) Neutral	&	0.31	&	0.18	&	0.23	&	22\\
		(2) Positive & 0.96	&	0.96	&	0.96	&	4215\\
		Average / Total	&	0.93	&	0.93	&	0.93	&	4526\\
		\bottomrule
		\end{tabular}\\
		\label{table:statistical-report-sentiment}
\end{table}

Table \ref{table:statistical-report-sentiment} corroborates our findings on biased classification towards the class with highest frequency distribution, supported by the confusion matrix in Figure \ref{conf_matrix_sentiment} (where $0$ represents the \textit{negative} class, $1$ represents the \textit{neutral} class, and $2$ represents the \textit{positive} class). We can see in this report that the model had a relatively weaker predictive performance for the negative and neutral sentiments.

The empirical evidences presented in this paper indicates a relatively high-performing predictive performance on both recommendation classification and sentiment classification, and this is despite the imbalanced class frequency distribution in the dataset. Such result supports the claim that using Bidirectional RNN-LSTM better captures the context of review texts which leads to better predictive performance. However, to further substantiate this claim, we recommend employing a uni-directional RNN-LSTM on the same classification problems for fair comparison.

\section{Conclusion and Recommendation}

To further improve the model, hyper-parameter tuning must be performed. This study was limited to an arbitrarily-chosen hyper-parameters due to computational cost restrictions. In addition, $k$-fold cross validation may give us a better and/or additional insight on the predictive performance of the model.\\
\indent Despite the limitations on the experiment for this study, it may be inferred that the Bidirectional RNN-LSTM model exhibited high performance (with F1-score of 0.88 for recommendation classification, and 0.93 for sentiment classification). Furthermore, the statistical measures on the classification problem may also be deemed satisfactory.

\section{Acknowledgment}
A sincere appreciation is given to Nick Brooks for his dataset on \textit{Women's Clothing E-Commerce Reviews}\cite{brooks2018women}, and also for granting us the permission to utilize some of his scripts for data visualizations\cite{brooks2018guided}.

\bibliographystyle{ACM-Reference-Format}
\bibliography{paper} 

\cleardoublepage
    \appendix
\chapter{Appendix}

\begin{figure*}[!htb]
\minipage{\textwidth}
\centering
	\includegraphics[width=\linewidth]{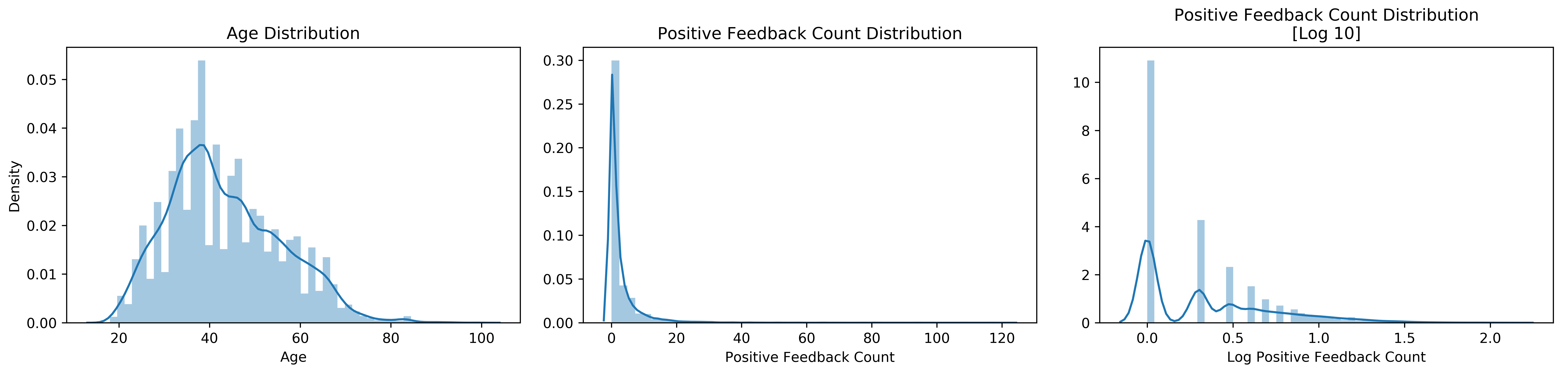}
	\caption{Plot generated based on script by \cite{brooks2018guided}. The frequency distribution of customer age and positive feedback.}
	\label{age-distribution}
\endminipage\hfill
\end{figure*}

\begin{figure*}[!htb]
\minipage{\textwidth}
\centering
	\includegraphics[width=\textwidth]{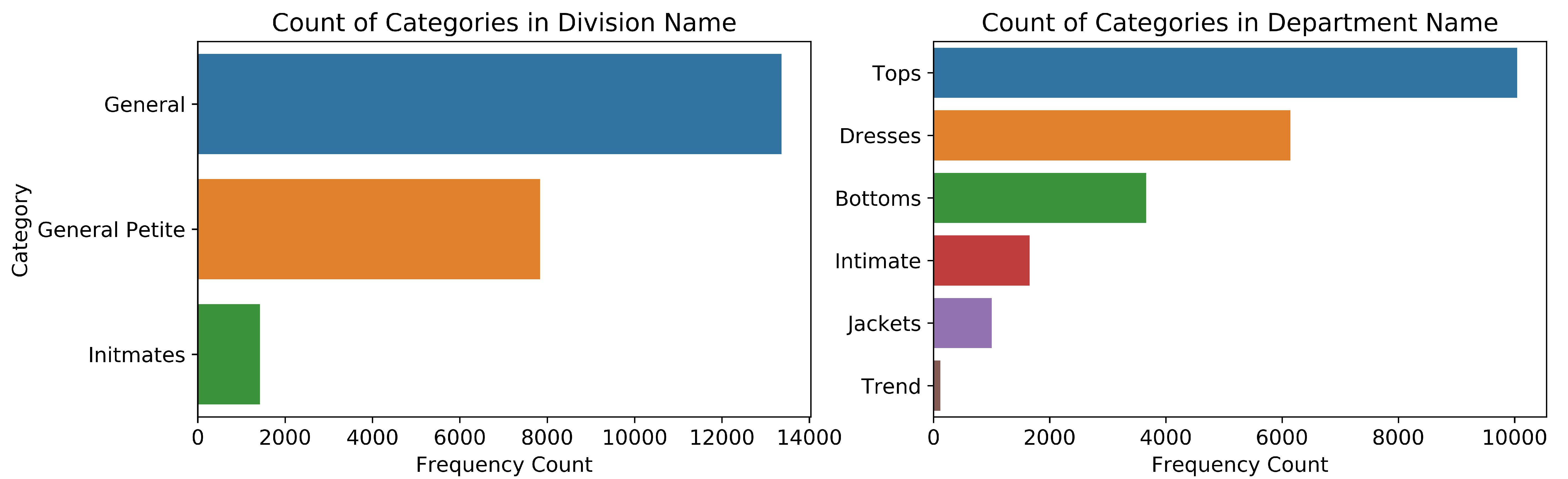}
	\caption{Plot generated based on script by \cite{brooks2018guided}. The frequency distribution of apparels per \textit{division} and \textit{department}.}
	\label{divname-and-deptname-freqdist}
\endminipage\hfill
\end{figure*}

\begin{figure*}[!htb]
\minipage{\textwidth}
\centering
	\includegraphics[width=\textwidth]{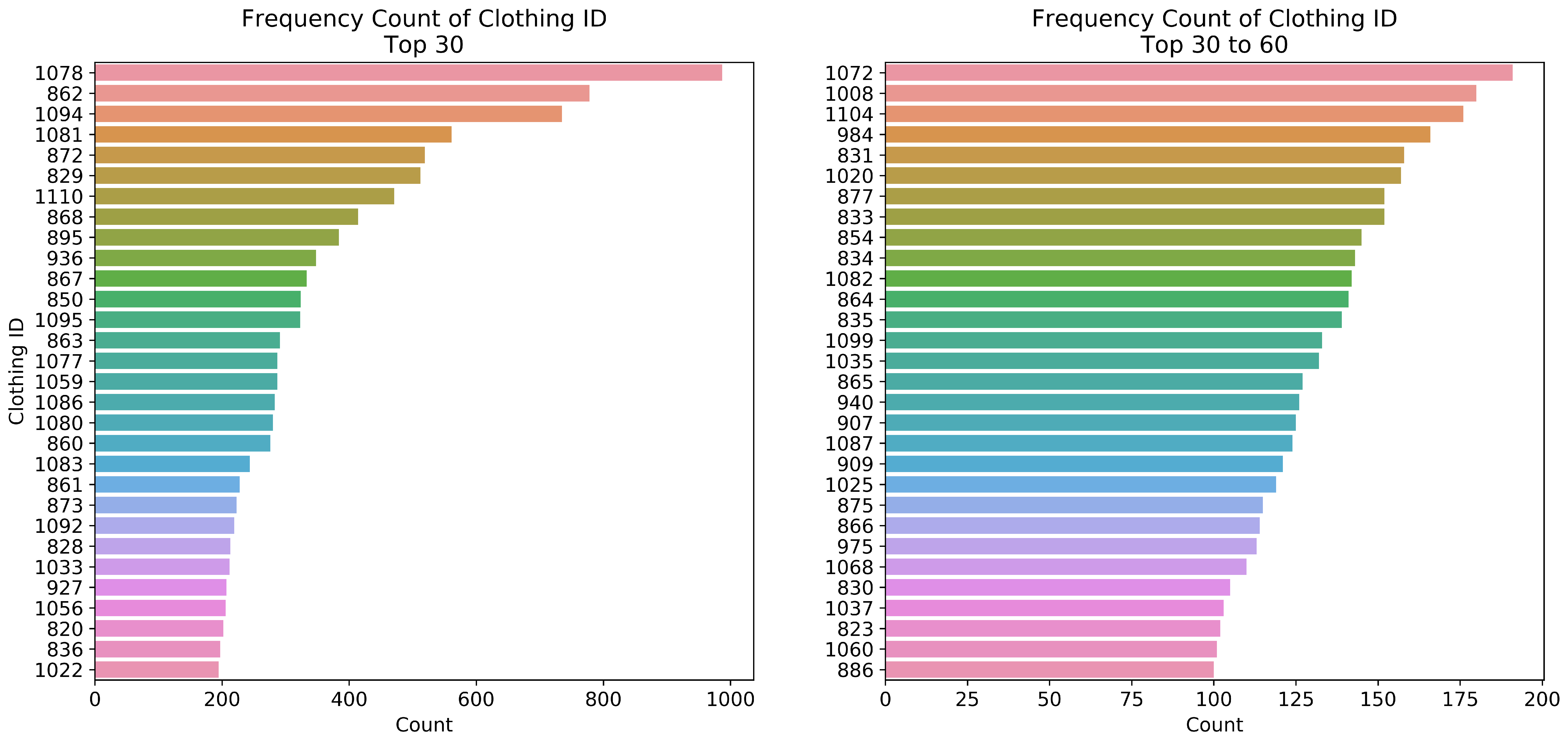}
	\caption{Plot generated based on script by \cite{brooks2018guided}. The frequency distribution of top 60 apparels per \textit{clothing ID}.}
	\label{top-60}
\endminipage\hfill
\end{figure*}

\begin{figure*}[!htb]
\minipage{0.75\textwidth}
\centering
	\includegraphics[width=\textwidth]{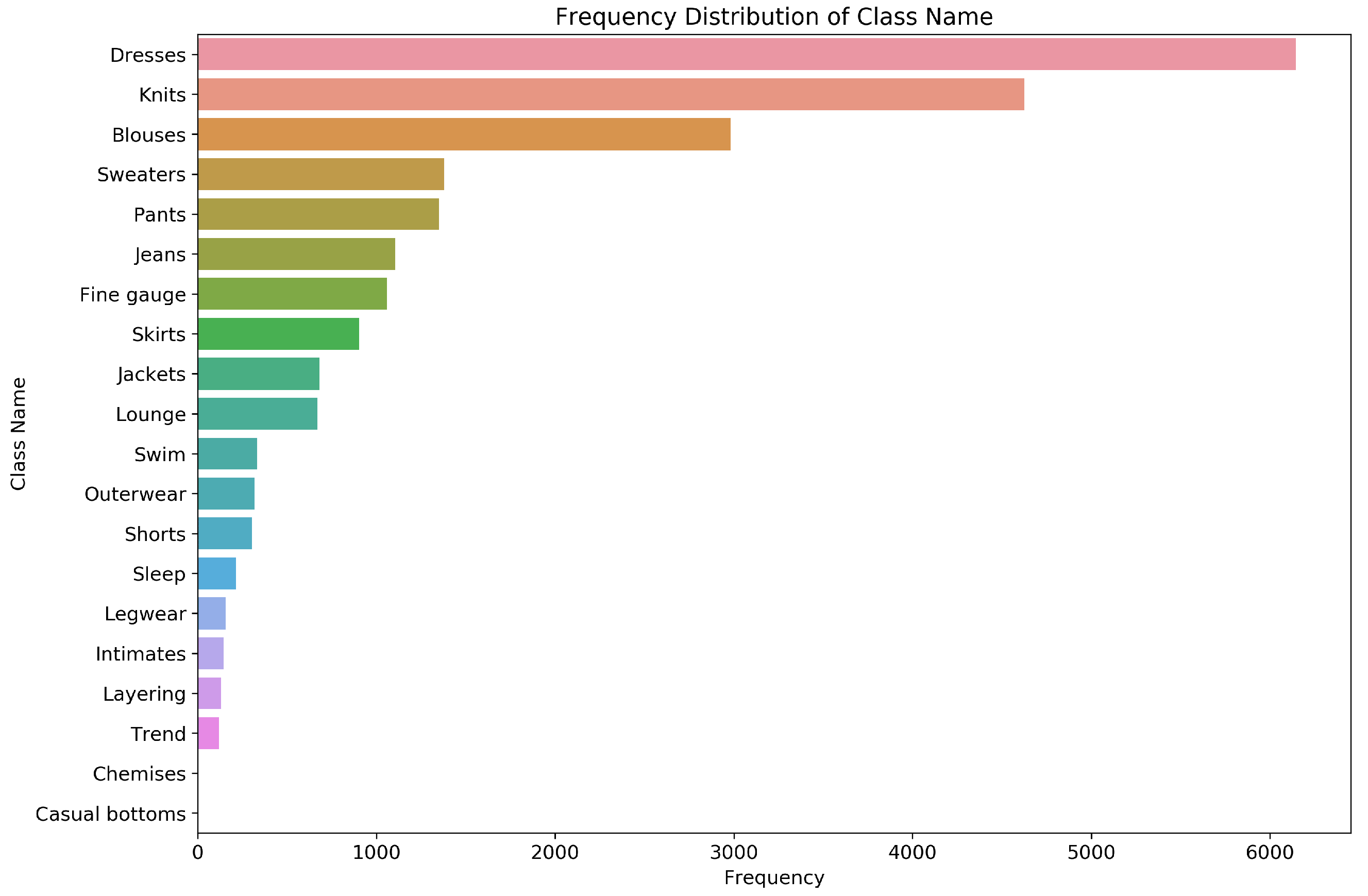}
	\caption{Plot generated based on script by \cite{brooks2018guided}. The frequency distribution of apparels per \textit{class}.}
	\label{classname-freqdist}
\endminipage\hfill
\end{figure*}

\begin{figure*}[!htb]
\minipage{\textwidth}
\centering
	\includegraphics[width=\textwidth]{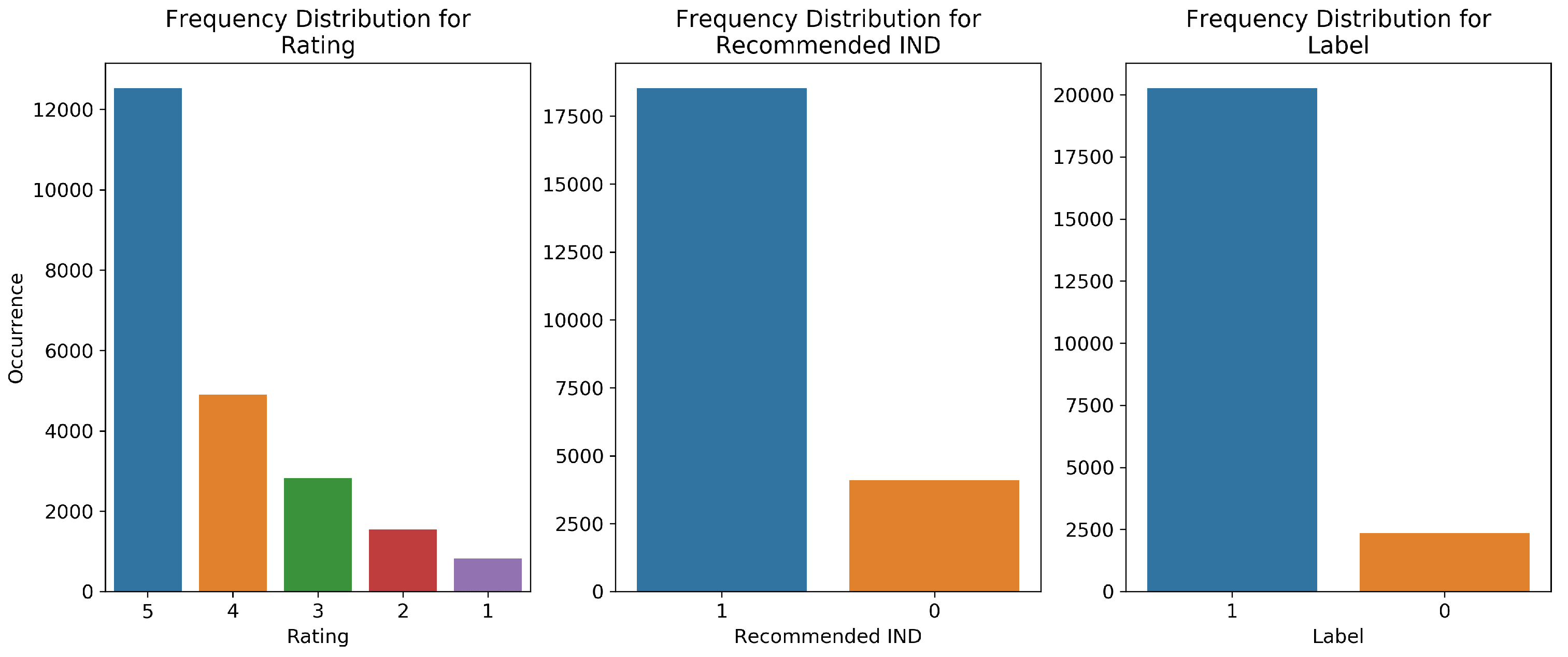}
	\caption{Plot generated based on script by \cite{brooks2018guided}. The frequency distribution of review ratings, recommendation, and labels.}
	\label{rating-recommended-label}
\endminipage\hfill
\end{figure*}

\begin{figure*}[!htb]
\minipage{\textwidth}
\centering
	\includegraphics[width=\textwidth]{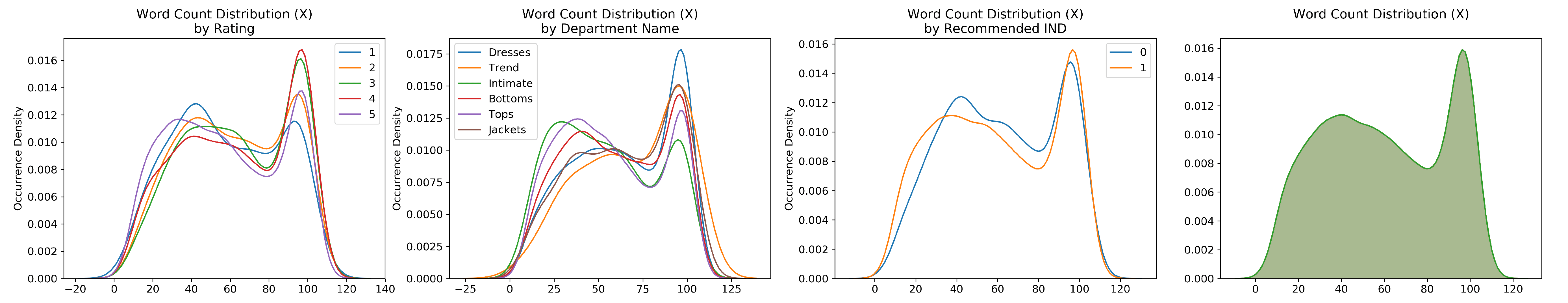}
	\caption{Plot generated based on script by \cite{brooks2018guided}. The word frequency distrubtion in review texts per rating, department, and recommendation.}
	\label{word-length}
\endminipage\hfill
\end{figure*}

\begin{figure*}[htb!]
\minipage{\textwidth}
\centering
	\includegraphics[width=\textwidth]{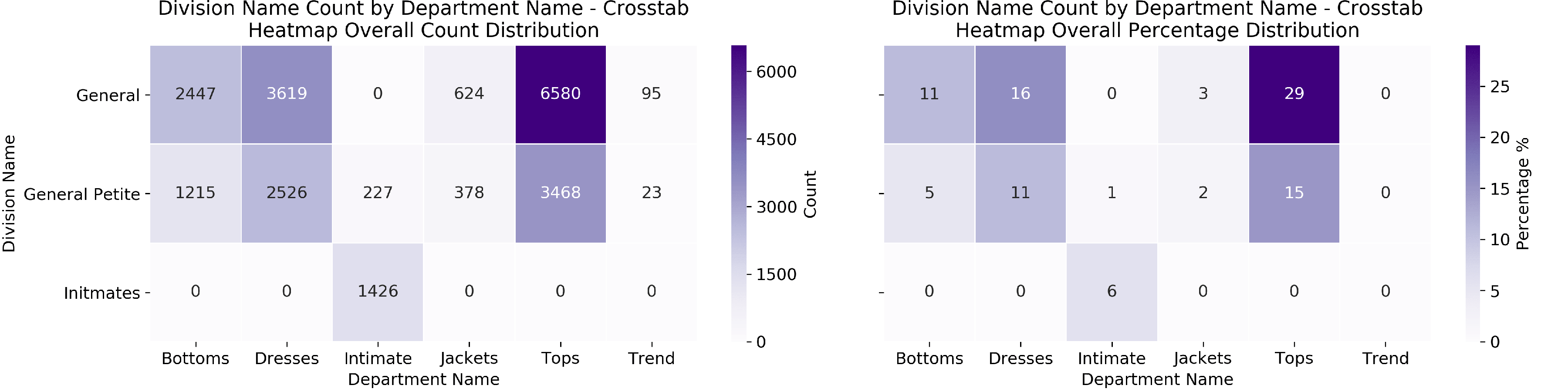}
	\caption{Heatmap generated based on script by \cite{brooks2018guided}. The cross tabulation for apparel per \textit{division} and \textit{department}.}
	\label{divname-deptname}
\endminipage\hfill
\end{figure*}

\begin{figure*}[htb!]
\minipage{\textwidth}
\centering
	\includegraphics[width=\textwidth]{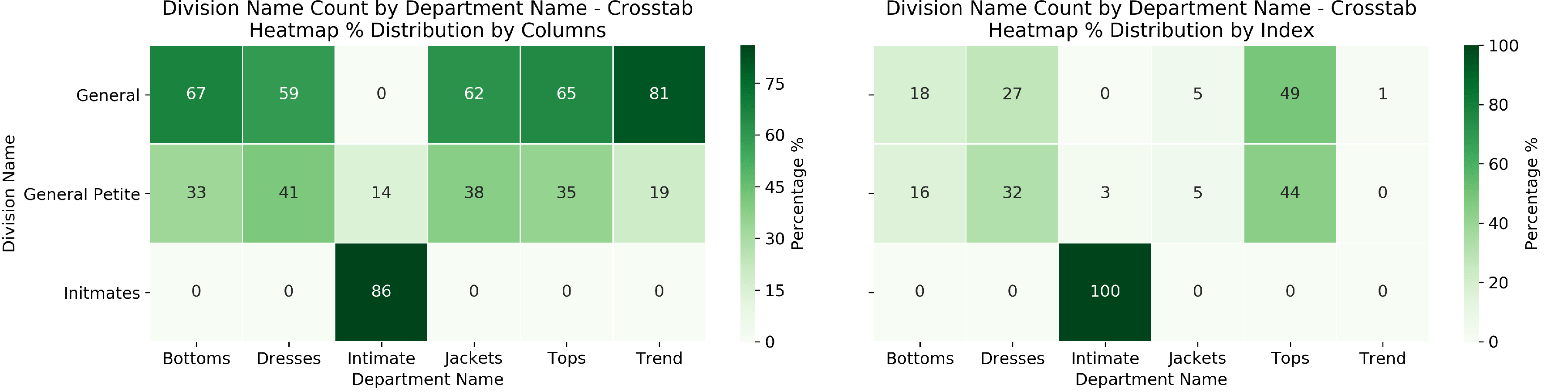}
	\caption{Heatmap generated based on script by \cite{brooks2018guided}. The normalized cross tabulation for apparel per \textit{division} and \textit{department}.}
	\label{divname-deptname-pivot}
\endminipage\hfill
\end{figure*}

\begin{figure*}[htb!]
\minipage{\textwidth}
\centering
	\includegraphics[width=\textwidth]{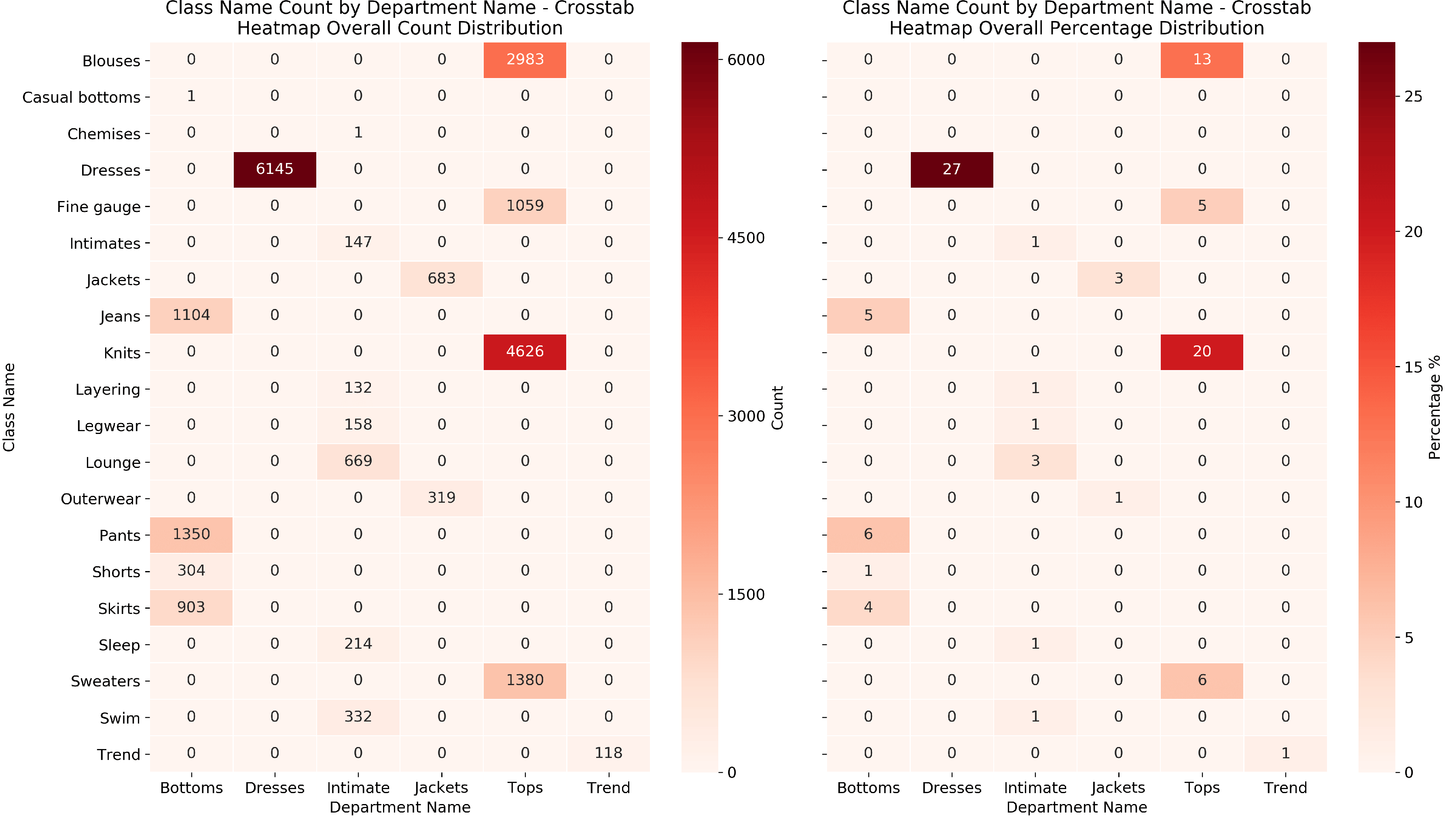}
	\caption{Heatmap generated based on script by \cite{brooks2018guided}. The cross tabulation for apparel per \textit{class} and \textit{department}.}
	\label{classname-deptname}
\endminipage\hfill
\end{figure*}

\begin{figure*}[htb!]
\minipage{\textwidth}
\centering
	\includegraphics[width=\textwidth]{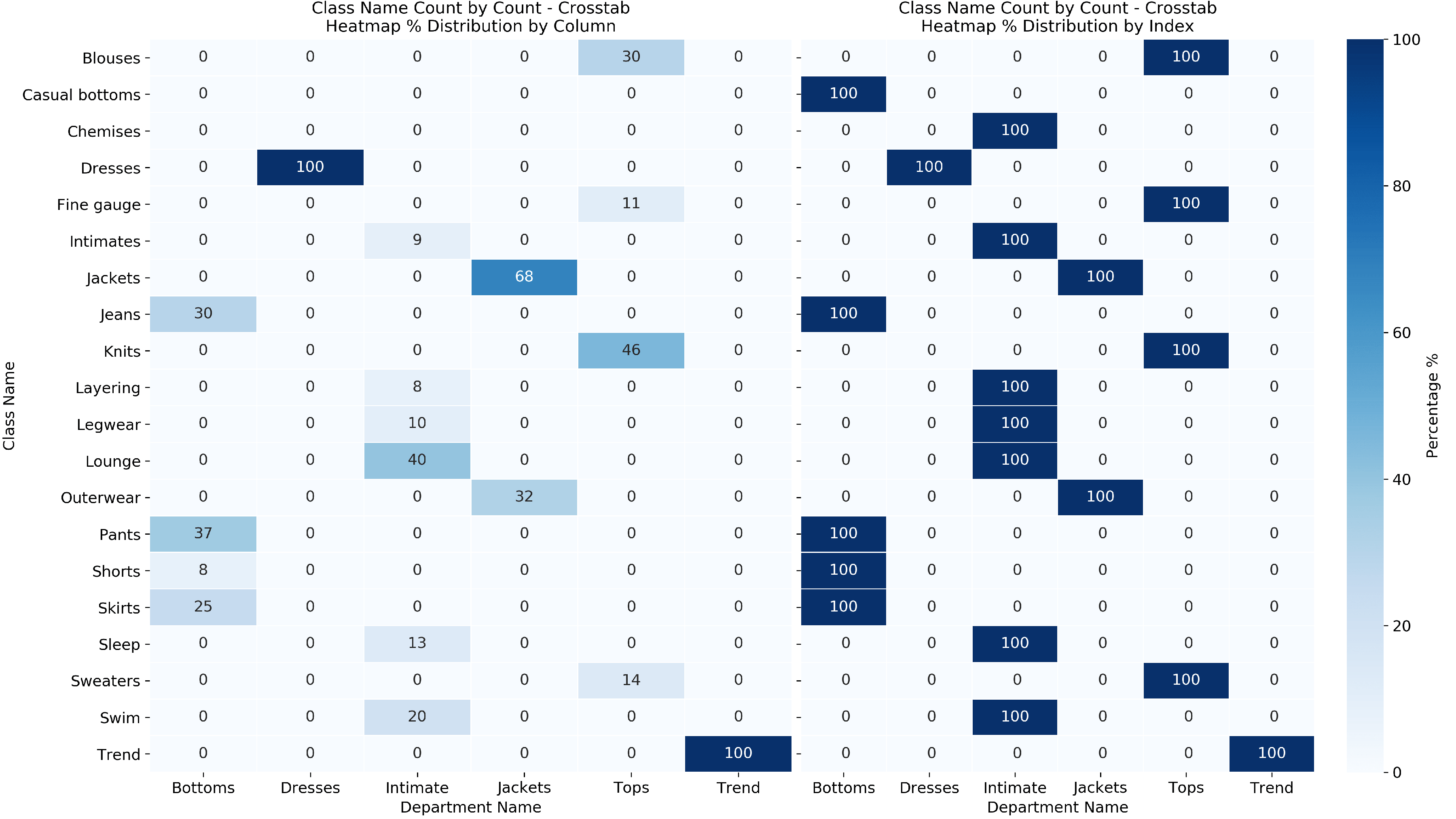}
	\caption{Heatmap generated based on script by \cite{brooks2018guided}. The normalized cross tabulation for apparel per \textit{class} and \textit{department}.}
	\label{classname-deptname-pivot}
\endminipage\hfill
\end{figure*}

\begin{figure*}[htb!]
\minipage{\textwidth}
\centering
	\includegraphics[width=\textwidth]{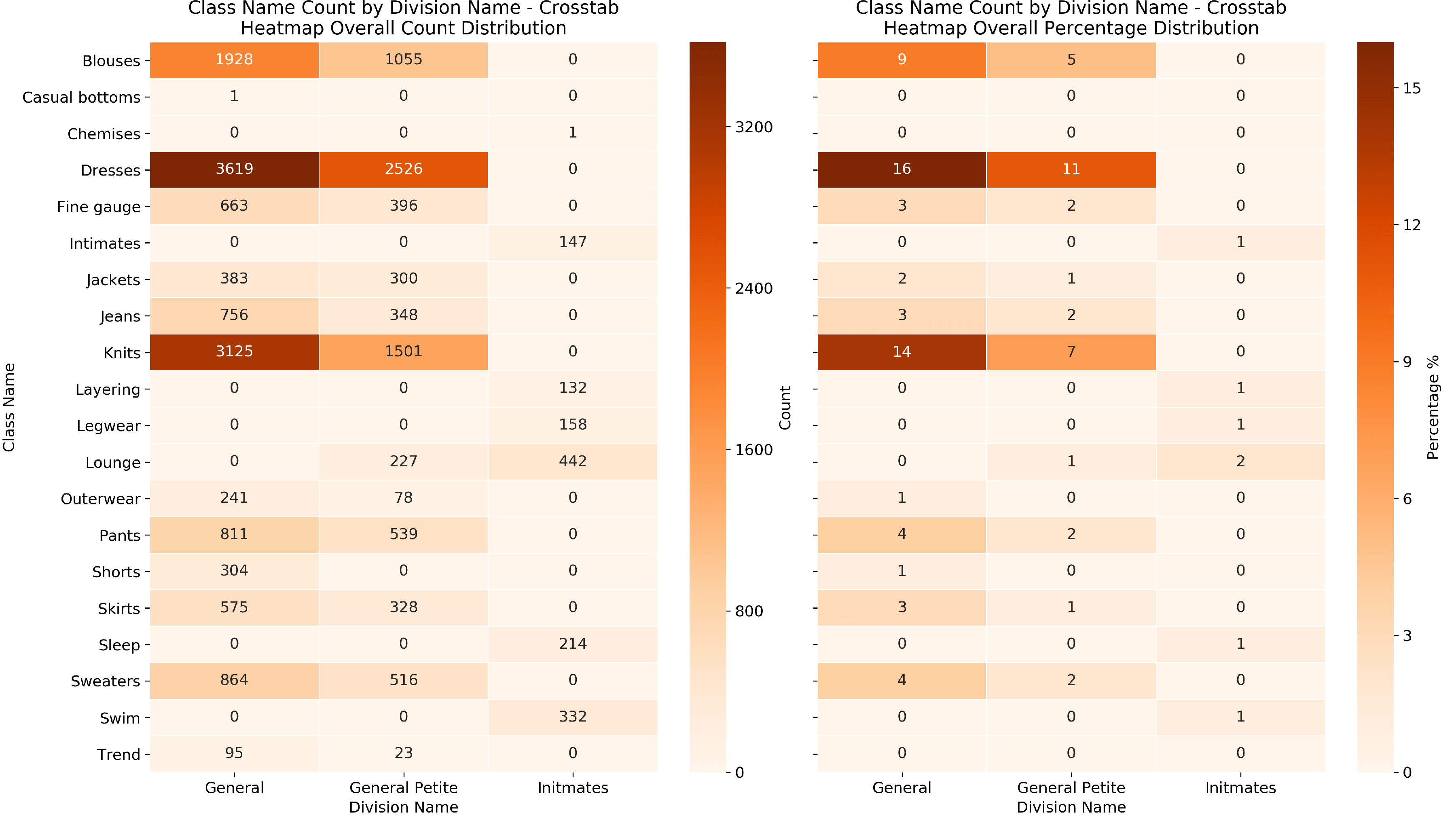}
	\caption{Heatmap generated based on script by \cite{brooks2018guided}. The cross tabulation for apparel per \textit{class} and \textit{division}.}
	\label{classname-divname}
\endminipage\hfill
\end{figure*}

\begin{figure*}[htb!]
\minipage{\textwidth}
\centering
	\includegraphics[width=\textwidth]{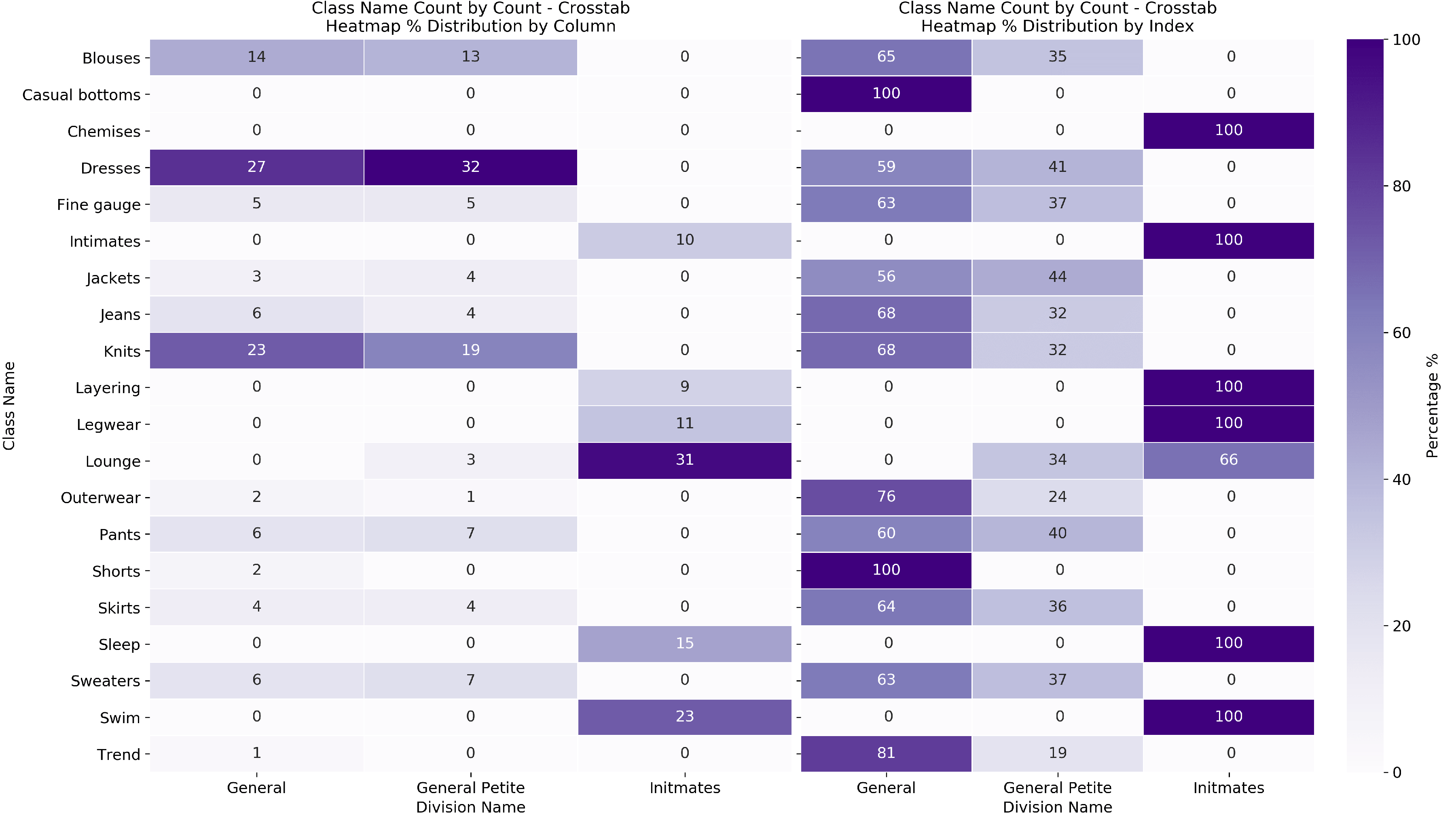}
	\caption{Heatmap generated based on script by \cite{brooks2018guided}. The normalized cross tabulation for apparel per \textit{class} and \textit{division}.}
	\label{classname-divname-pivot}
\endminipage\hfill
\end{figure*}

\begin{figure*}[htb!]
\minipage{0.5\textwidth}
\centering
	\includegraphics[width=\textwidth]{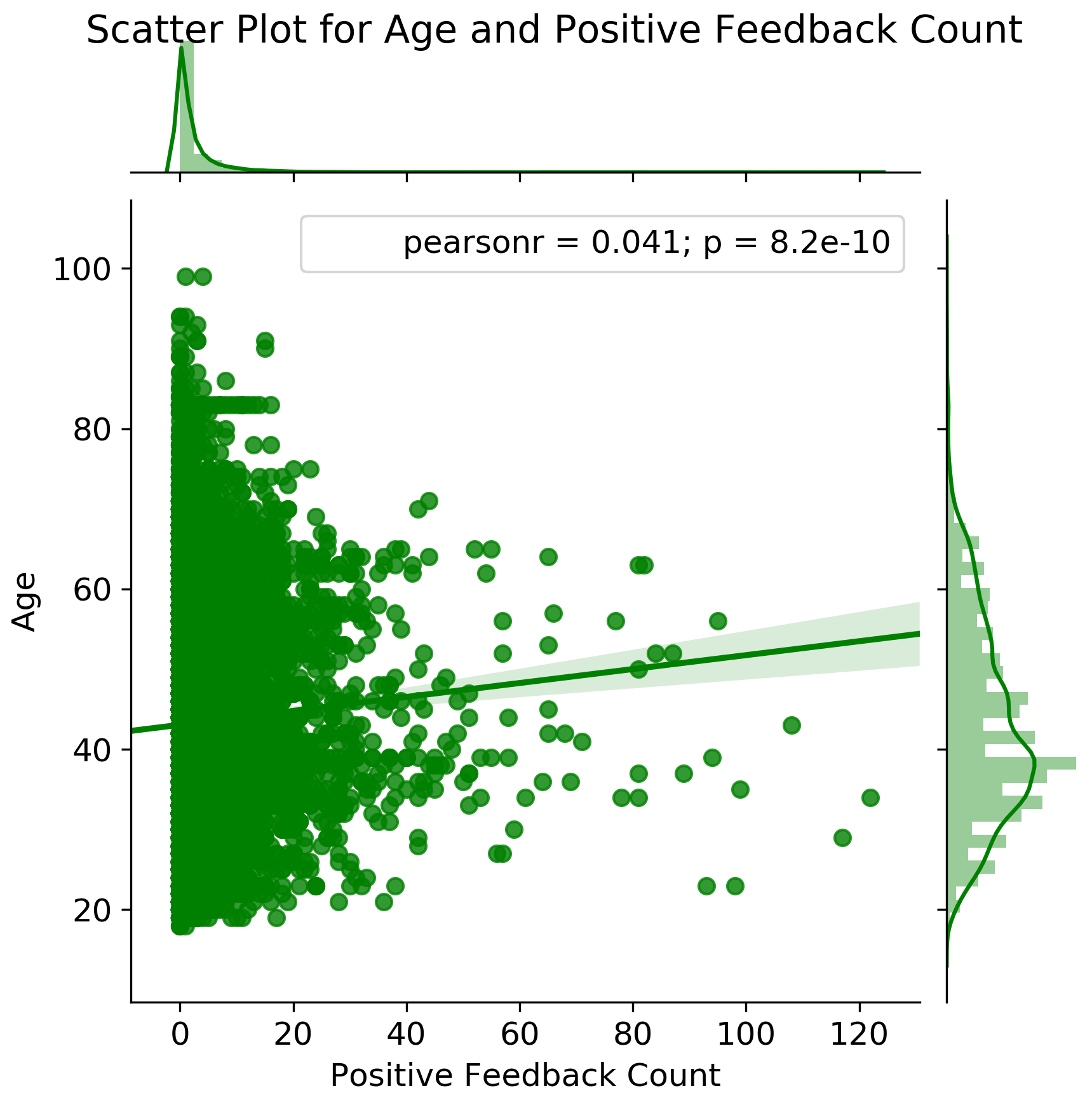}
	\caption{Scatter plot generated based on script by \cite{brooks2018guided}. The scatter plot for age and positive feedback count.}
	\label{age-positivefeedback-scatter}
\endminipage\hfill
\end{figure*}

\begin{figure*}[htb!]
\minipage{\textwidth}
\centering
	\includegraphics[width=\textwidth]{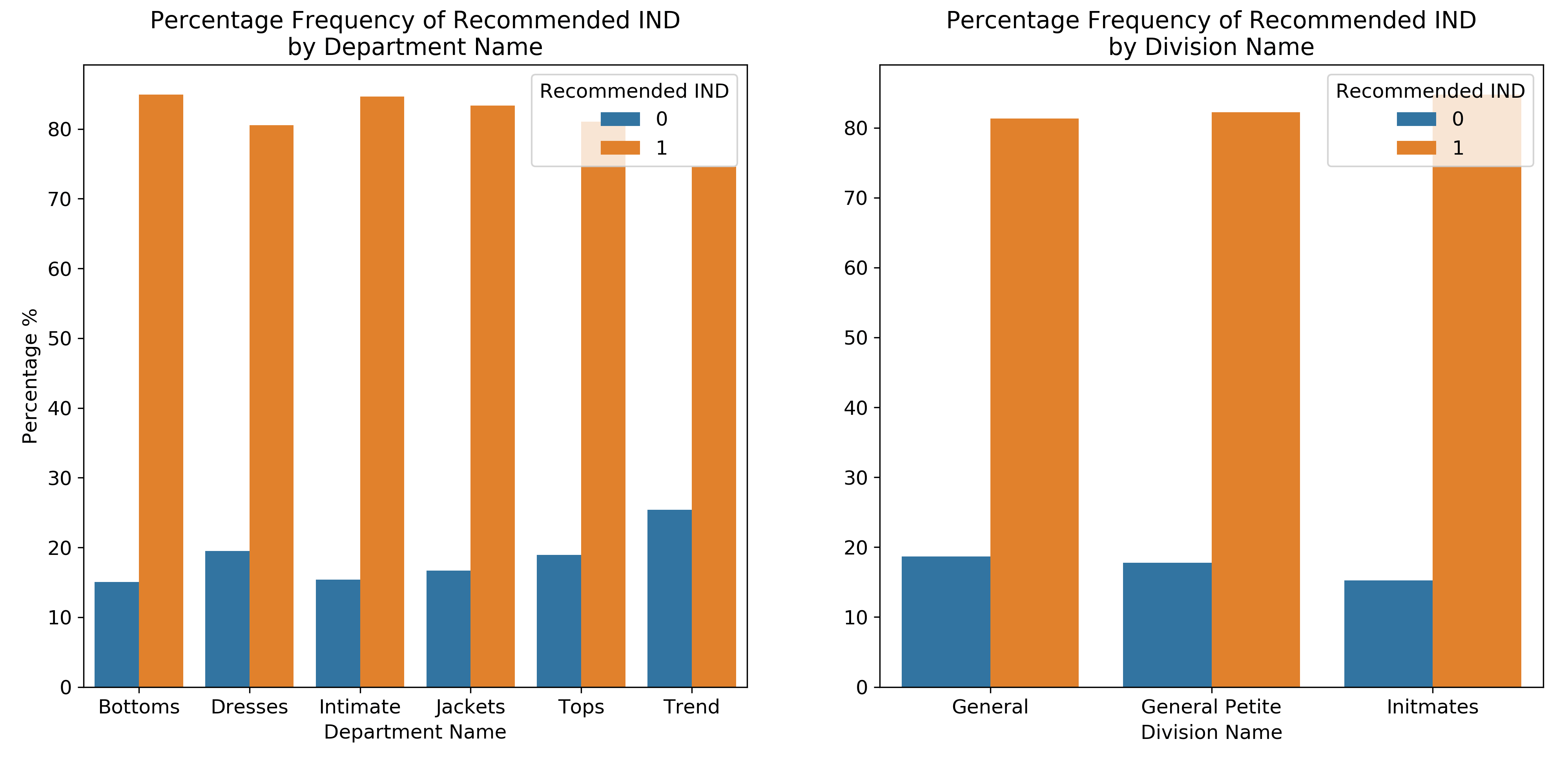}
	\caption{Plot generated based on script by \cite{brooks2018guided}. The percentage frequency of recommendation indicator per review by \textit{department} and \textit{division}.}
	\label{recommendation-deptname-divname}
\endminipage\hfill
\end{figure*}

\begin{figure*}[htb!]
\minipage{\textwidth}
\centering
	\includegraphics[width=\textwidth]{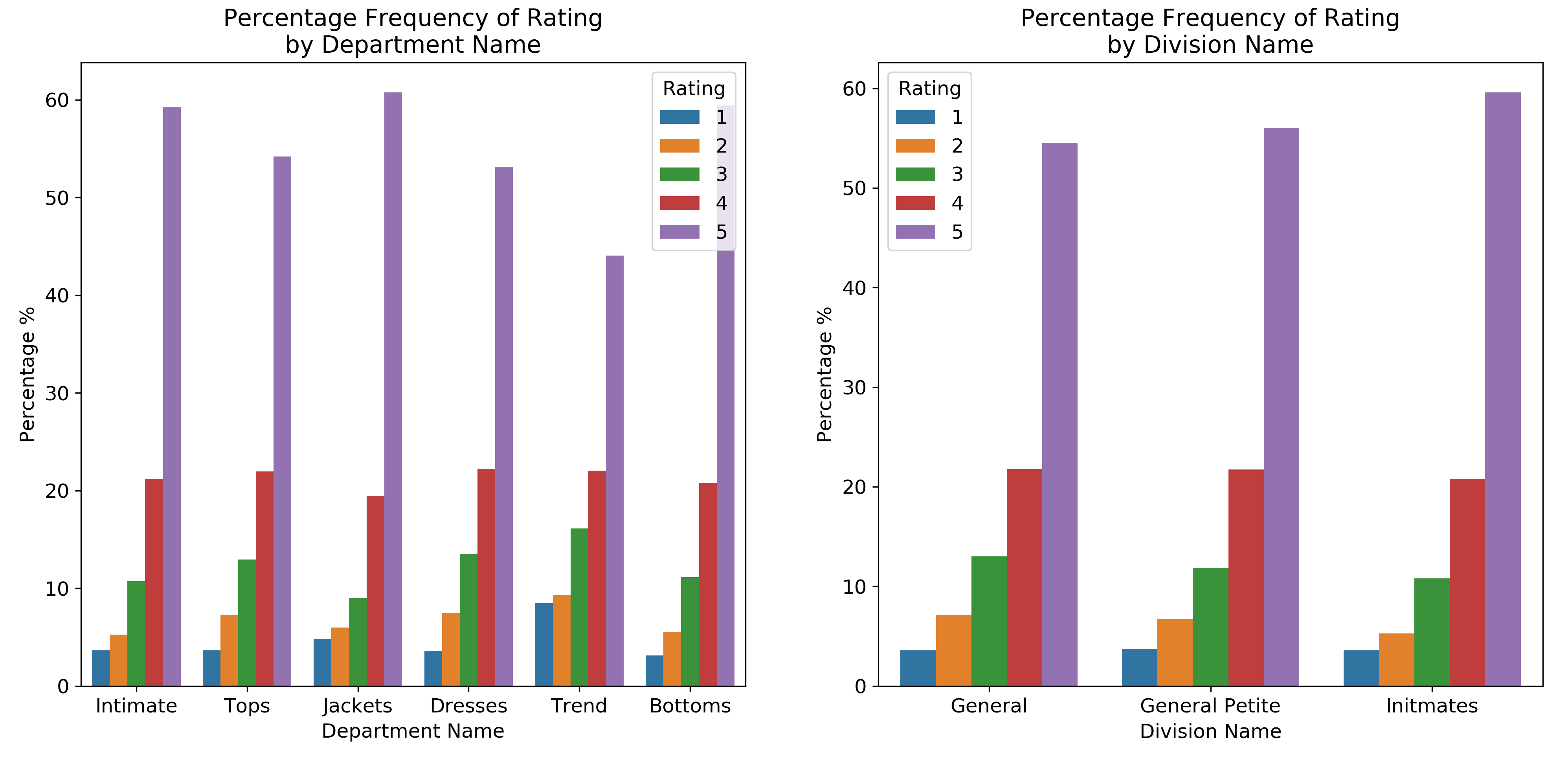}
	\caption{Plot generated based on script by \cite{brooks2018guided}. The percentage frequency of review rating by \textit{department} and \textit{division}.}
	\label{rating-deptname-divname}
\endminipage\hfill
\end{figure*}

\begin{figure*}[htb!]
\minipage{\textwidth}
\centering
	\includegraphics[width=\textwidth]{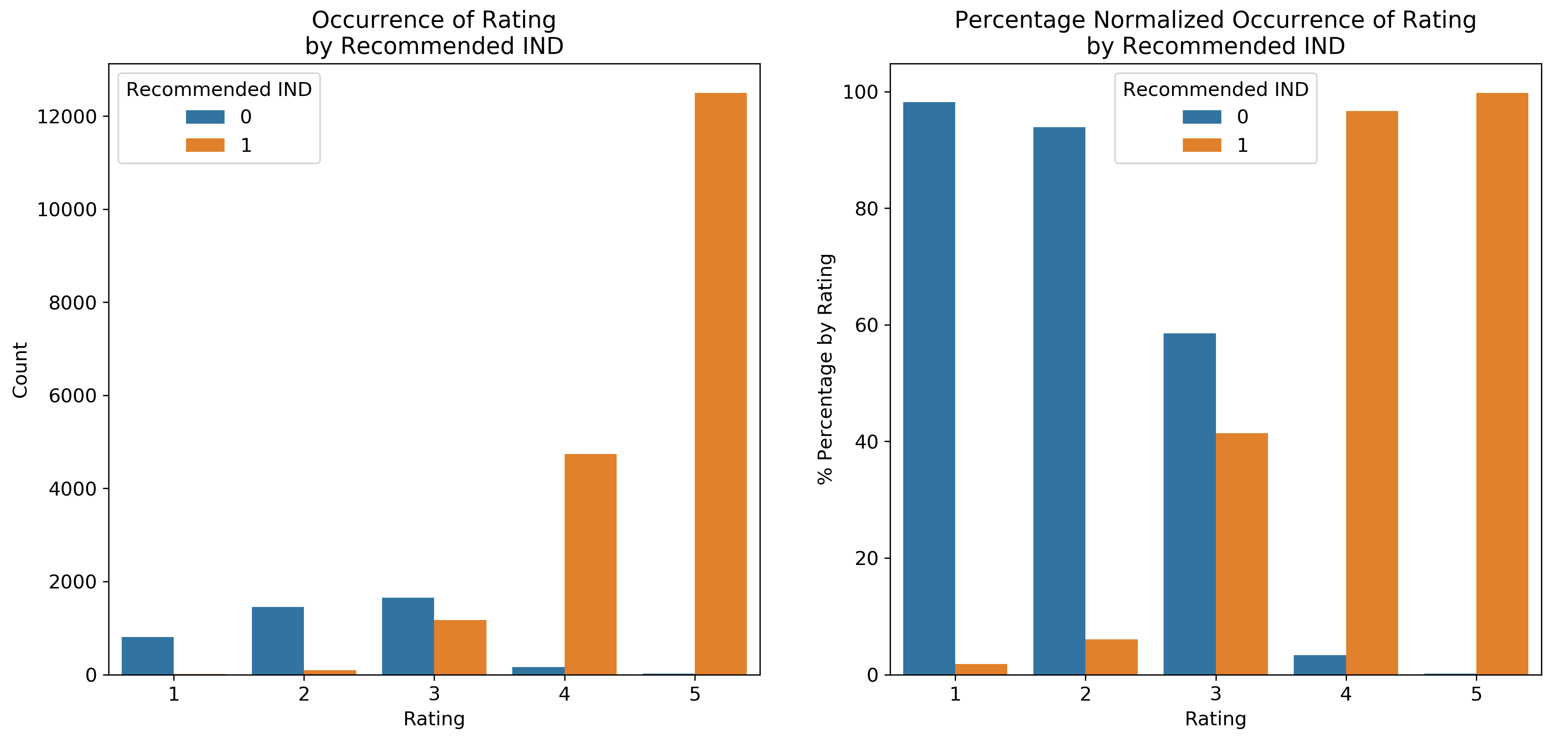}
	\caption{Plot generated based on script by \cite{brooks2018guided}. The frequency of rating by recommendation indicator.}
	\label{rating-recommended}
\endminipage\hfill
\end{figure*}

\begin{figure*}[!htb]
\minipage{\textwidth}
\centering
	\includegraphics[width=\textwidth]{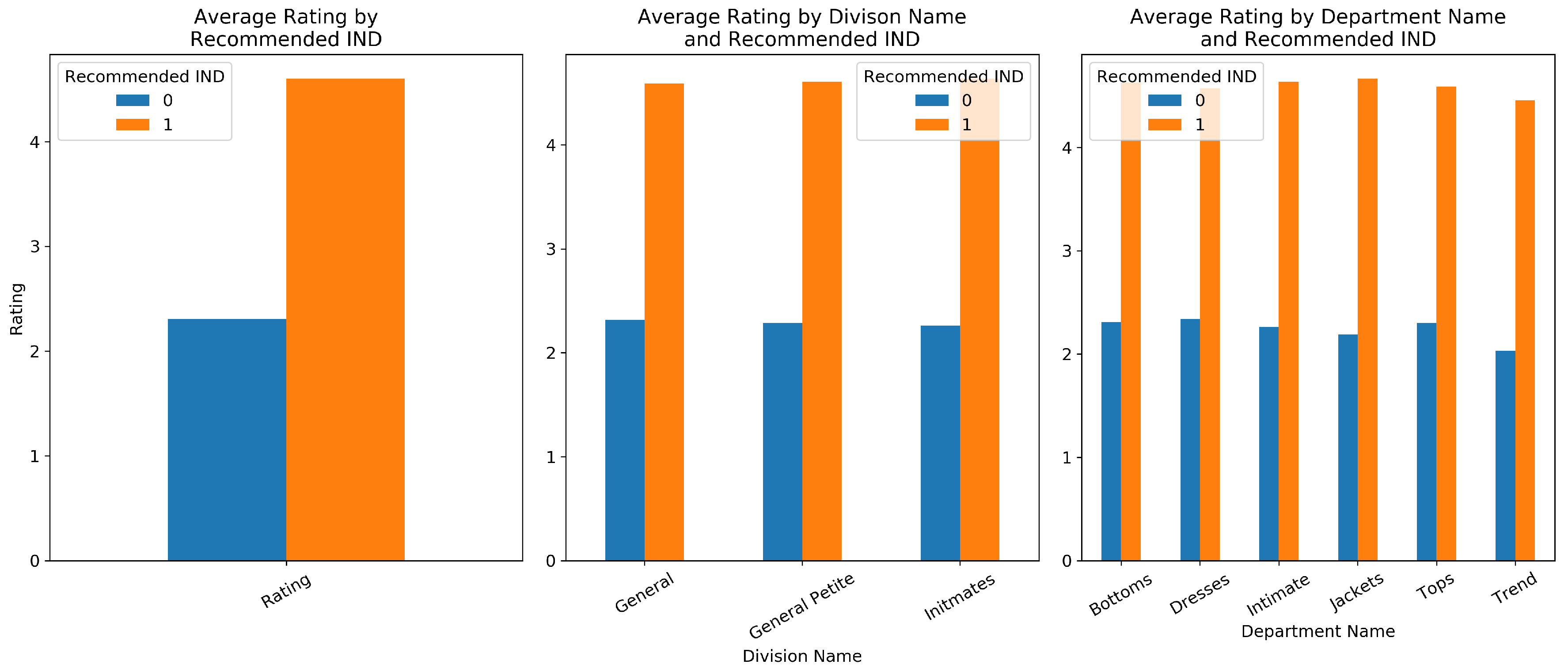}
	\caption{Plot generated based on script by \cite{brooks2018guided}. The average rating frequency by \textit{division} \textit{department}, and recommendation indicator.}
	\label{averagerating-deptname-recommended}
\endminipage\hfill
\end{figure*}

\begin{figure*}[!htb]
\minipage{0.75\textwidth}
\centering
	\includegraphics[width=\textwidth]{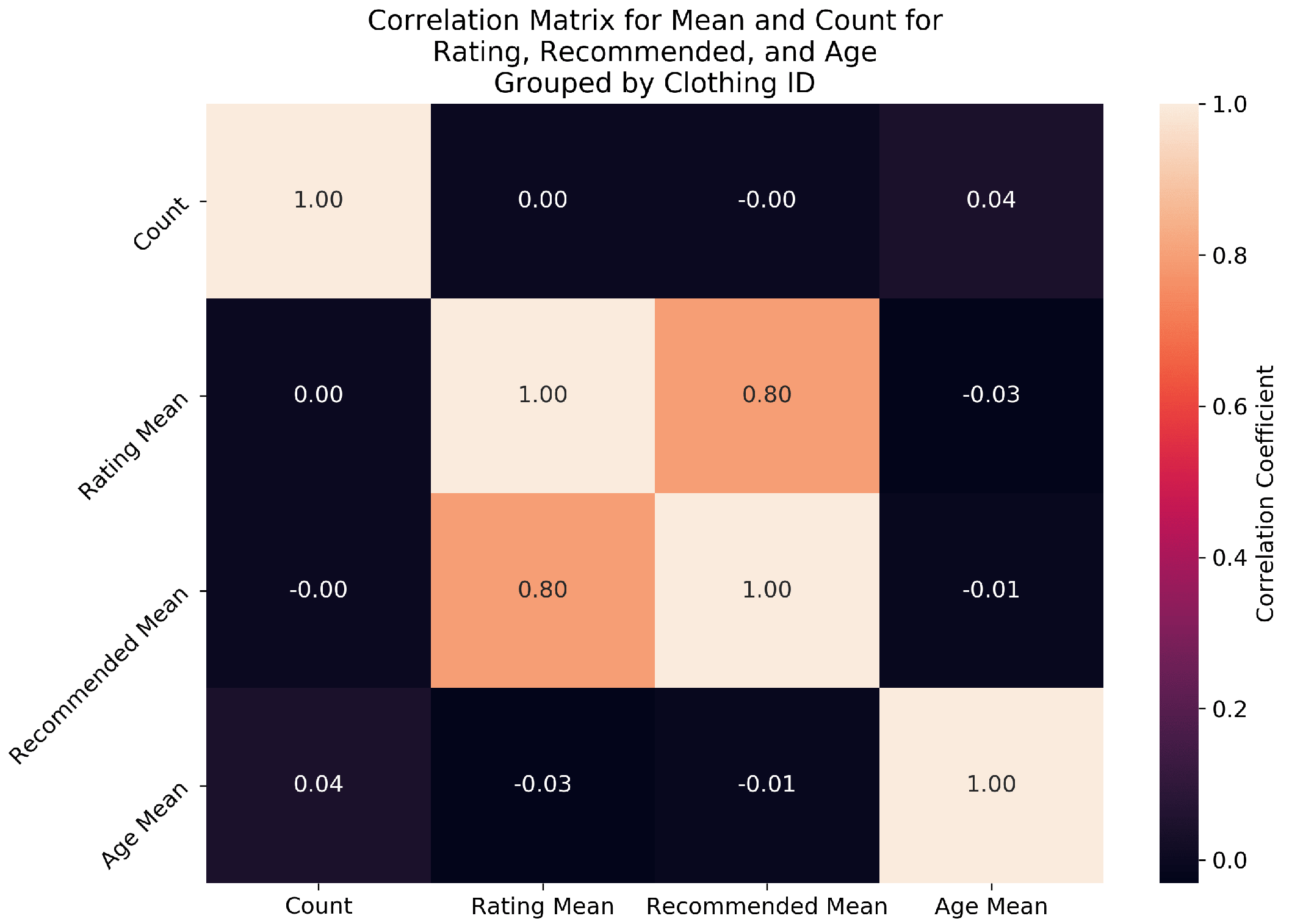}
	\caption{Heatmap generated based on script by \cite{brooks2018guided}. The correlation matrix for average rating and recommendation indicator, grouped by clothing ID.}
	\label{meanrating-recommended-clothing-corr}
\endminipage\hfill
\end{figure*}

\begin{figure*}[!htb]
\minipage{\textwidth}
\centering
	\includegraphics[width=\textwidth]{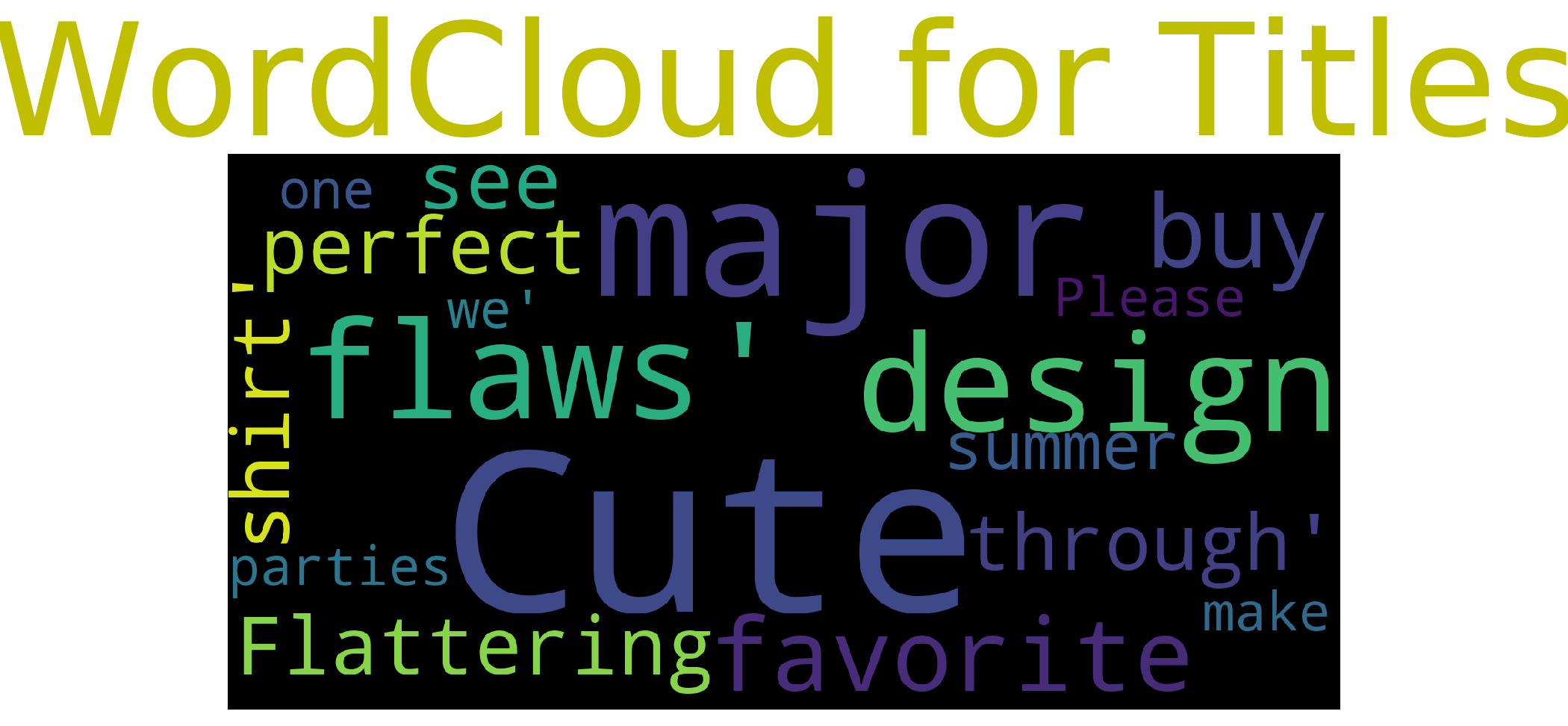}
	\caption{Word cloud generated based on script by \cite{brooks2018guided}. The most frequent words used for review titles.}
	\label{wordcloud-titles}
\endminipage\hfill
\end{figure*}

\begin{figure*}[!htb]
\minipage{\textwidth}
\centering
	\includegraphics[width=\textwidth]{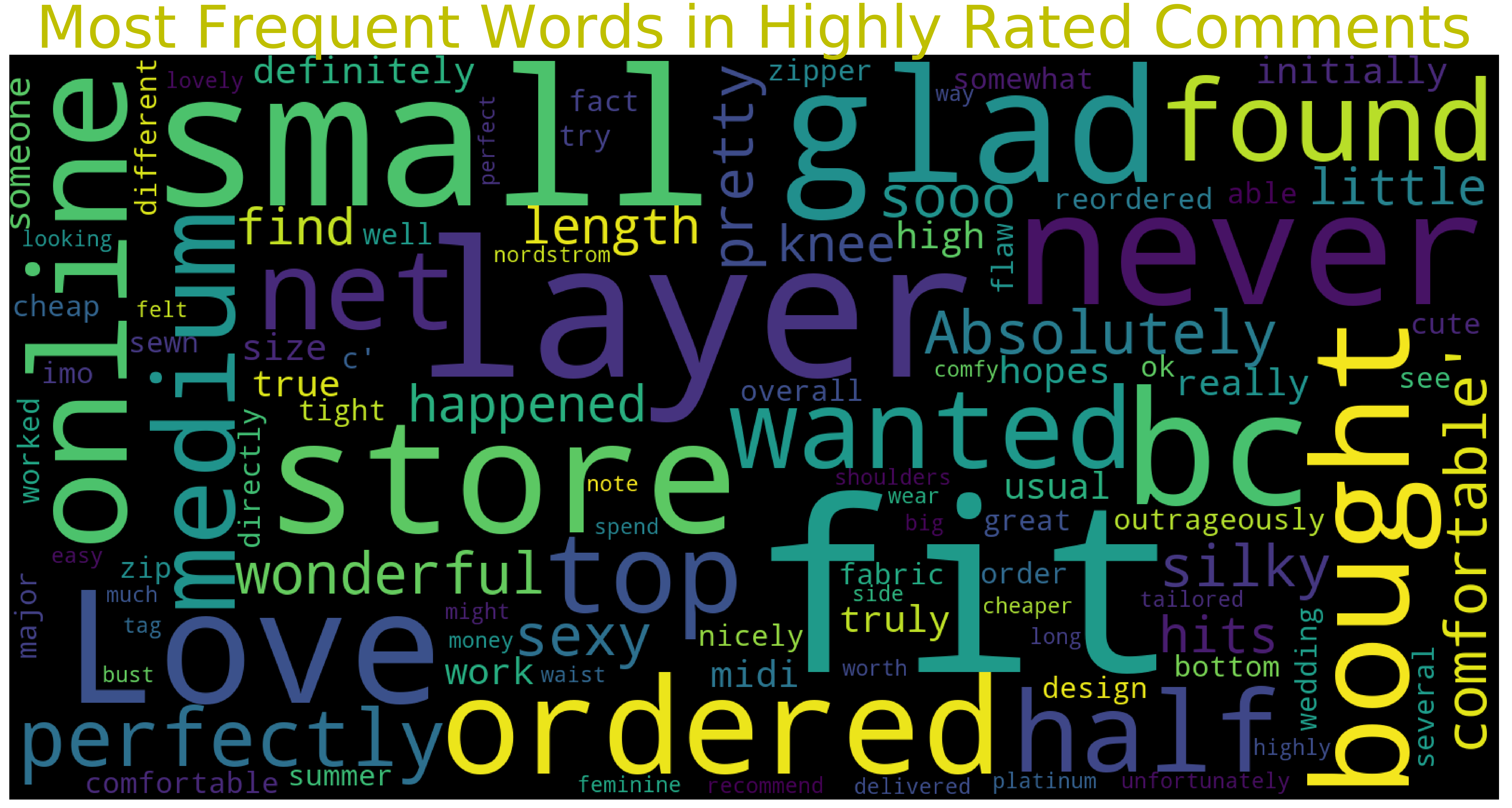}
	\caption{Word cloud generated based on script by \cite{brooks2018guided}. The most frequent words used in review texts with high ratings.}
	\label{wc-most-freq-words-high-rate-comments}
\endminipage\hfill
\end{figure*}

\begin{figure*}[!htb]
\minipage{\textwidth}
\centering
	\includegraphics[width=\textwidth]{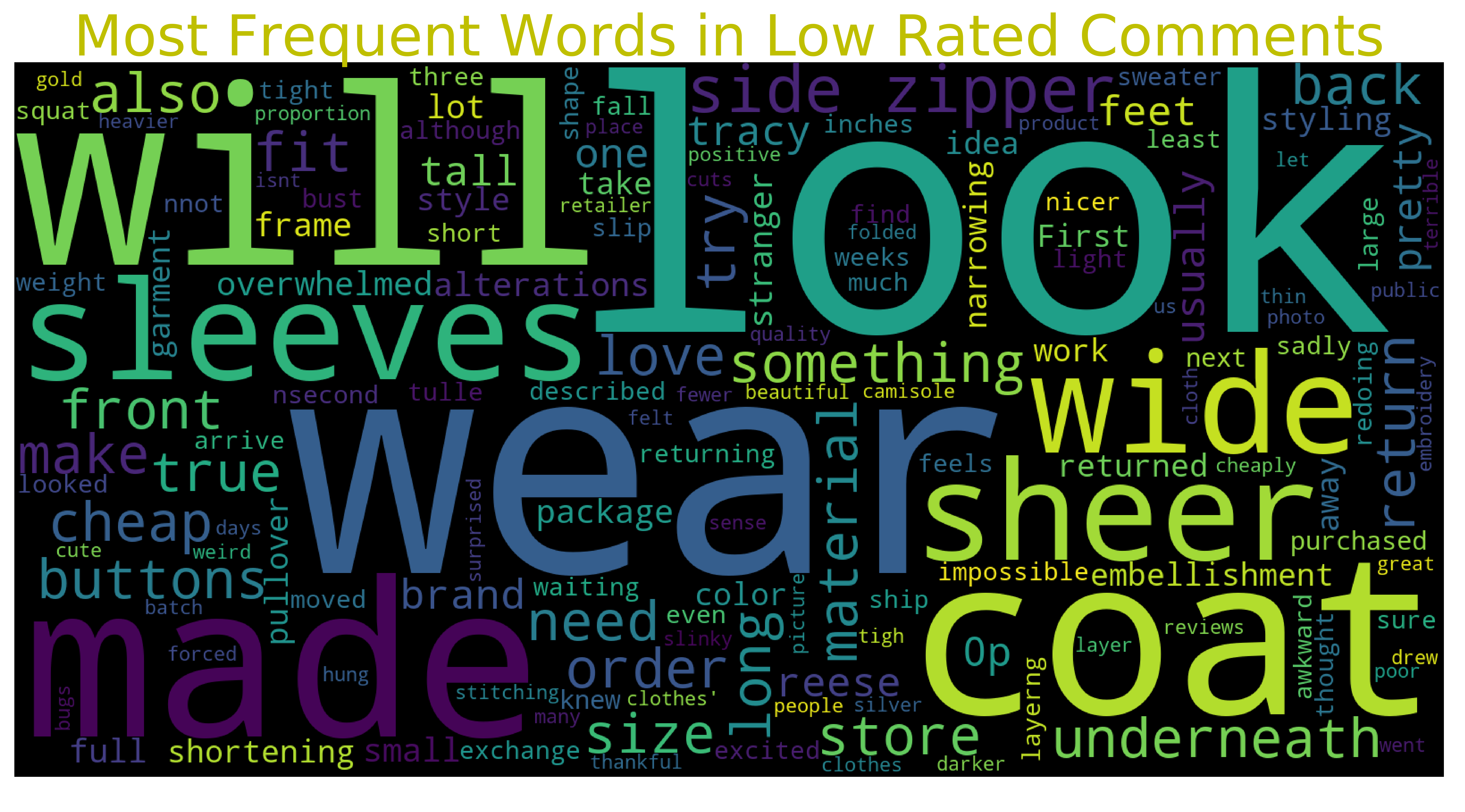}
	\caption{Word cloud generated based on script by \cite{brooks2018guided}. The most frequent words used in review texts with low ratings.}
	\label{most-freq-words-low-rate-comment}
\endminipage\hfill
\end{figure*}

\begin{figure*}[!htb]
\minipage{\textwidth}
\centering
	\includegraphics[width=\textwidth]{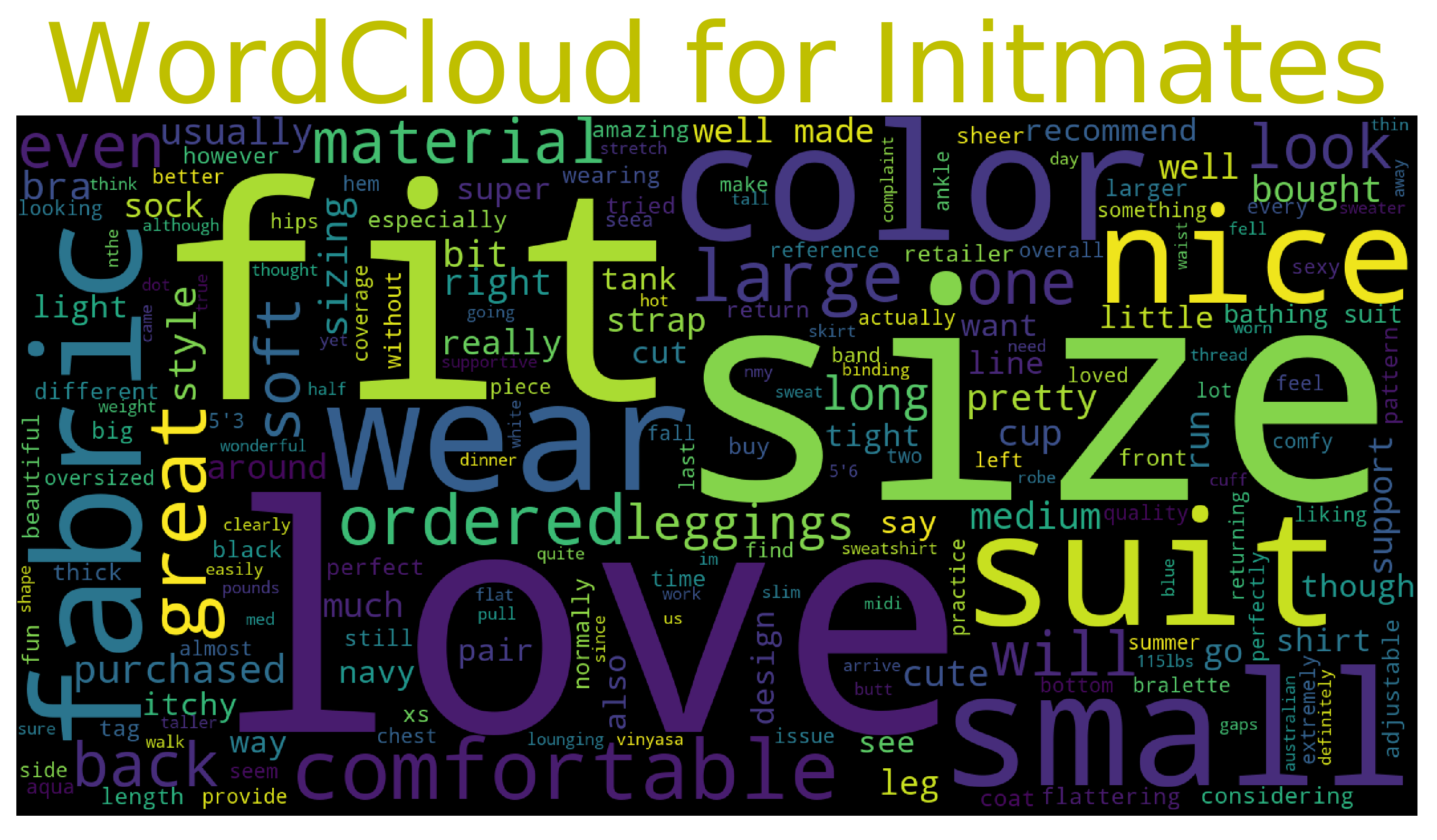}
	\caption{Word cloud generated based on script by \cite{brooks2018guided}. The most frequent words used in review texts in \textit{intimate} apparels.}
	\label{wordcloud-intimates}
\endminipage\hfill
\end{figure*}

\begin{figure*}[!htb]
\minipage{\textwidth}
\centering
	\includegraphics[width=\textwidth]{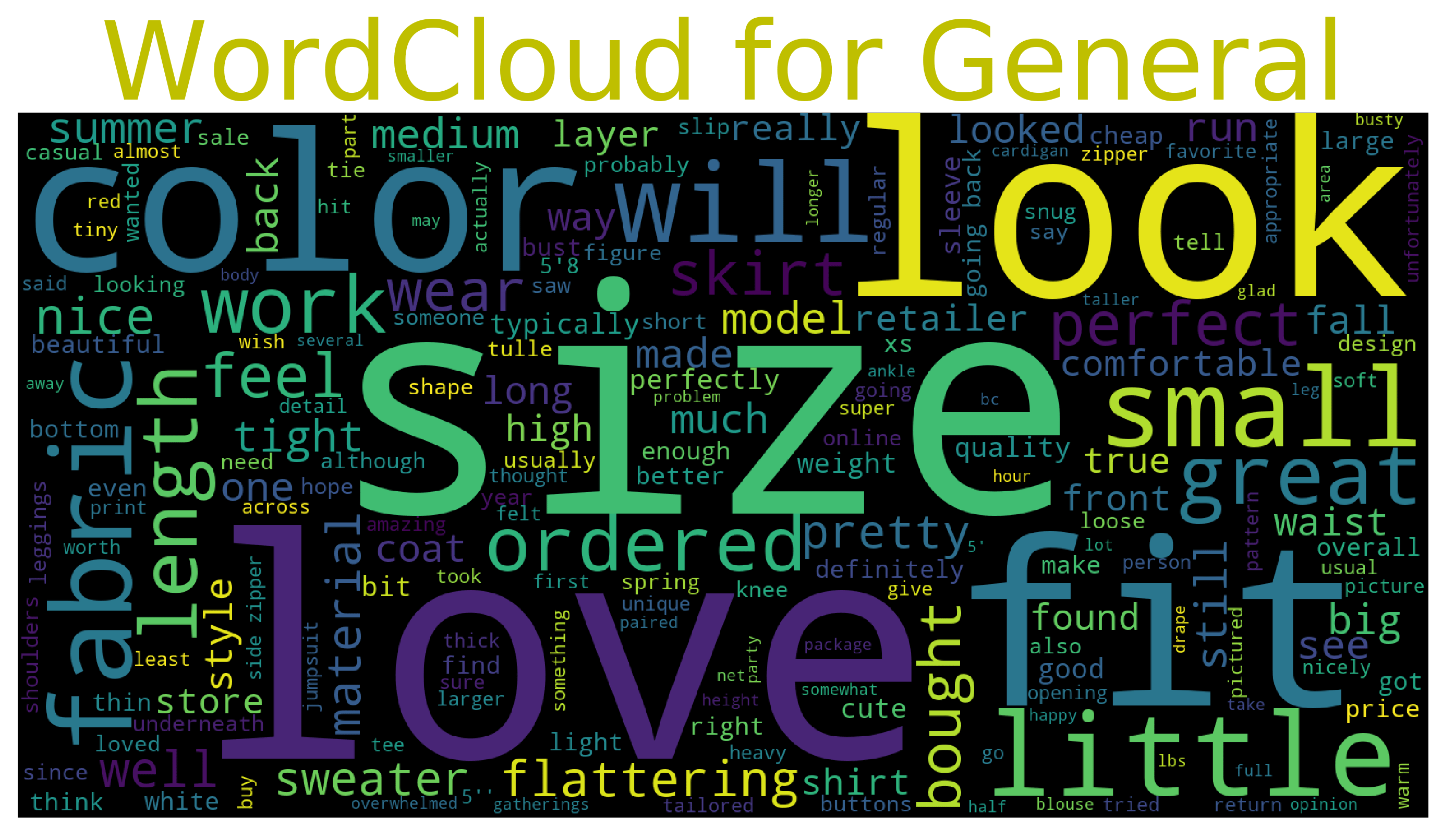}
	\caption{Word cloud generated based on script by \cite{brooks2018guided}. The most frequent words used in review texts in \textit{general-sized} apparels.}
	\label{wordcloud-general}
\endminipage\hfill
\end{figure*}

\begin{figure*}[!htb]
\minipage{\textwidth}
\centering
	\includegraphics[width=\textwidth]{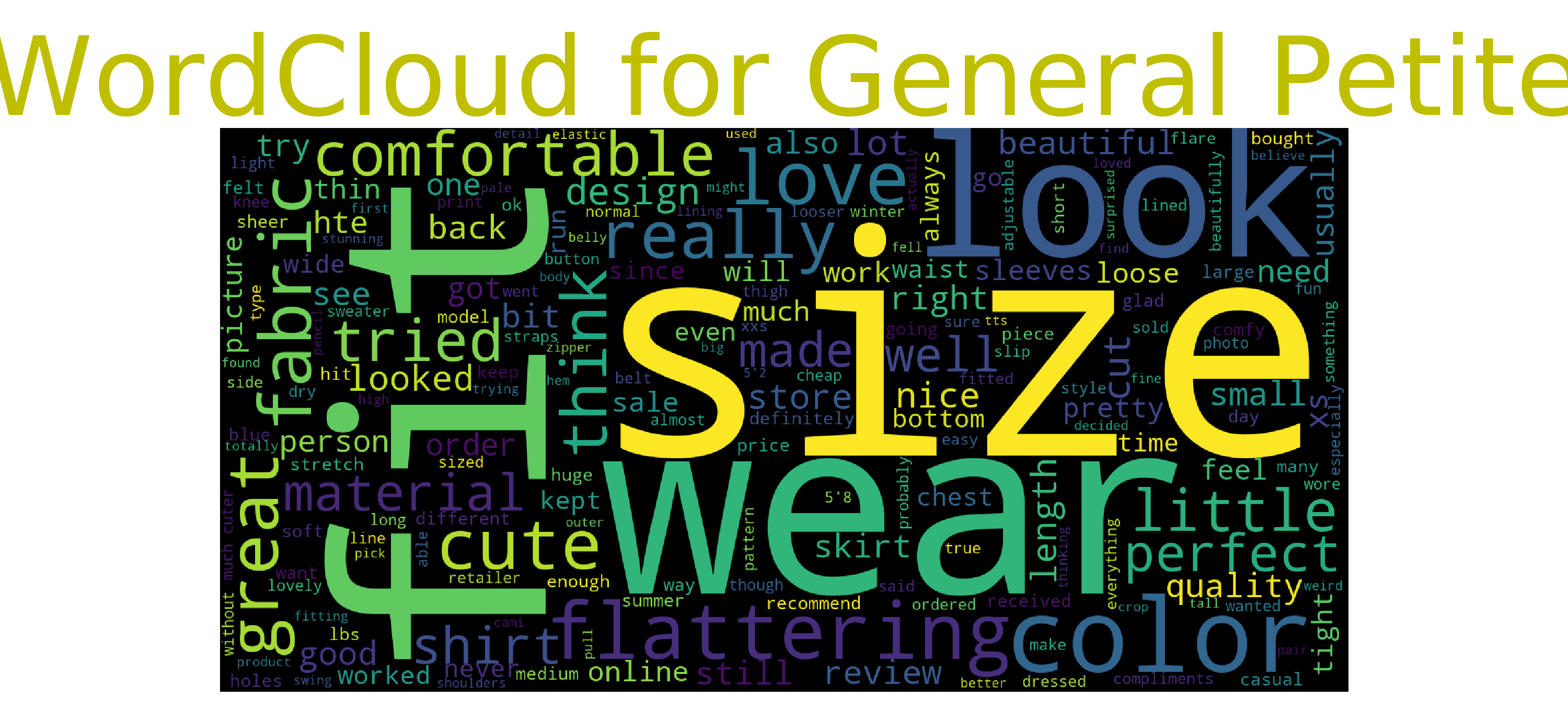}
	\caption{Word cloud generated based on script by \cite{brooks2018guided}. The most frequent words used in review texts in \textit{petite-sized} apparels.}
	\label{wordcloud-petite}
\endminipage\hfill
\end{figure*}

\begin{figure*}[!htb]
 \minipage{0.75\textwidth}
 \centering
 	\includegraphics[width=\textwidth]{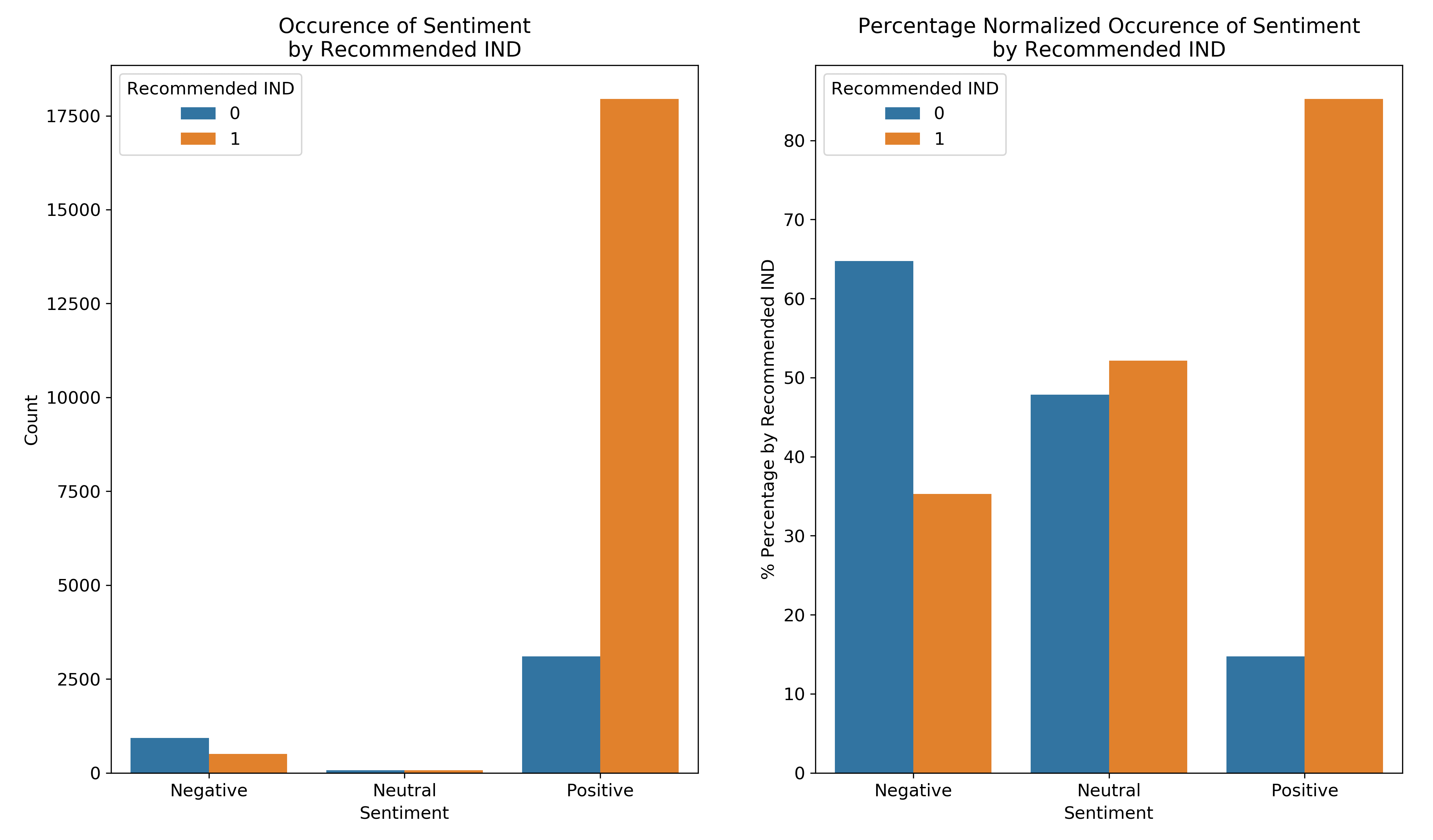}
 	\caption{Plot generated based on script by \cite{brooks2018guided}. The frequency distribution of sentiments per recommendation state in review texts.}
 	\label{norm-sentimentdist}
\endminipage\hfill
\end{figure*}

\cleardoublepage
    \listoffigures

%% file: paper.bbl

\begin{thebibliography}{14}


\ifx \showCODEN    \undefined \def \showCODEN     #1{\unskip}     \fi
\ifx \showDOI      \undefined \def \showDOI       #1{#1}\fi
\ifx \showISBNx    \undefined \def \showISBNx     #1{\unskip}     \fi
\ifx \showISBNxiii \undefined \def \showISBNxiii  #1{\unskip}     \fi
\ifx \showISSN     \undefined \def \showISSN      #1{\unskip}     \fi
\ifx \showLCCN     \undefined \def \showLCCN      #1{\unskip}     \fi
\ifx \shownote     \undefined \def \shownote      #1{#1}          \fi
\ifx \showarticletitle \undefined \def \showarticletitle #1{#1}   \fi
\ifx \showURL      \undefined \def \showURL       {\relax}        \fi
\providecommand\bibfield[2]{#2}
\providecommand\bibinfo[2]{#2}
\providecommand\natexlab[1]{#1}
\providecommand\showeprint[2][]{arXiv:#2}

\bibitem[\protect\citeauthoryear{Abadi, Agarwal, Barham, Brevdo, Chen, Citro,
  Corrado, Davis, Dean, Devin, Ghemawat, Goodfellow, Harp, Irving, Isard, Jia,
  Jozefowicz, Kaiser, Kudlur, Levenberg, Man\'{e}, Monga, Moore, Murray, Olah,
  Schuster, Shlens, Steiner, Sutskever, Talwar, Tucker, Vanhoucke, Vasudevan,
  Vi\'{e}gas, Vinyals, Warden, Wattenberg, Wicke, Yu, and Zheng}{Abadi
  et~al\mbox{.}}{2015}]%
        {tensorflow2015-whitepaper}
\bibfield{author}{\bibinfo{person}{Mart\'{\i}n Abadi}, \bibinfo{person}{Ashish
  Agarwal}, \bibinfo{person}{Paul Barham}, \bibinfo{person}{Eugene Brevdo},
  \bibinfo{person}{Zhifeng Chen}, \bibinfo{person}{Craig Citro},
  \bibinfo{person}{Greg~S. Corrado}, \bibinfo{person}{Andy Davis},
  \bibinfo{person}{Jeffrey Dean}, \bibinfo{person}{Matthieu Devin},
  \bibinfo{person}{Sanjay Ghemawat}, \bibinfo{person}{Ian Goodfellow},
  \bibinfo{person}{Andrew Harp}, \bibinfo{person}{Geoffrey Irving},
  \bibinfo{person}{Michael Isard}, \bibinfo{person}{Yangqing Jia},
  \bibinfo{person}{Rafal Jozefowicz}, \bibinfo{person}{Lukasz Kaiser},
  \bibinfo{person}{Manjunath Kudlur}, \bibinfo{person}{Josh Levenberg},
  \bibinfo{person}{Dandelion Man\'{e}}, \bibinfo{person}{Rajat Monga},
  \bibinfo{person}{Sherry Moore}, \bibinfo{person}{Derek Murray},
  \bibinfo{person}{Chris Olah}, \bibinfo{person}{Mike Schuster},
  \bibinfo{person}{Jonathon Shlens}, \bibinfo{person}{Benoit Steiner},
  \bibinfo{person}{Ilya Sutskever}, \bibinfo{person}{Kunal Talwar},
  \bibinfo{person}{Paul Tucker}, \bibinfo{person}{Vincent Vanhoucke},
  \bibinfo{person}{Vijay Vasudevan}, \bibinfo{person}{Fernanda Vi\'{e}gas},
  \bibinfo{person}{Oriol Vinyals}, \bibinfo{person}{Pete Warden},
  \bibinfo{person}{Martin Wattenberg}, \bibinfo{person}{Martin Wicke},
  \bibinfo{person}{Yuan Yu}, {and} \bibinfo{person}{Xiaoqiang Zheng}.}
  \bibinfo{year}{2015}\natexlab{}.
\newblock \bibinfo{title}{{TensorFlow}: Large-Scale Machine Learning on
  Heterogeneous Systems}.
\newblock   (\bibinfo{year}{2015}).
\newblock
\showURL{%
\url{https://www.tensorflow.org/}}
\newblock
\shownote{Software available from tensorflow.org.}


\bibitem[\protect\citeauthoryear{Brooks}{Brooks}{2018a}]%
        {brooks2018guided}
\bibfield{author}{\bibinfo{person}{Nick Brooks}.}
  \bibinfo{year}{2018}\natexlab{a}.
\newblock \bibinfo{title}{Guided Numeric and Text Exploration E-Commerce}.
\newblock   (\bibinfo{year}{2018}).
\newblock
\showURL{%
\url{https://www.kaggle.com/nicapotato/guided-numeric-and-text-exploration-e-commerce}}


\bibitem[\protect\citeauthoryear{Brooks}{Brooks}{2018b}]%
        {brooks2018women}
\bibfield{author}{\bibinfo{person}{Nick Brooks}.}
  \bibinfo{year}{2018}\natexlab{b}.
\newblock \bibinfo{title}{Women's E-Commerce Clothing Reviews}.
\newblock   (\bibinfo{year}{2018}).
\newblock
\showURL{%
\url{https://www.kaggle.com/nicapotato/womens-ecommerce-clothing-reviews}}


\bibitem[\protect\citeauthoryear{Chollet et~al\mbox{.}}{Chollet
  et~al\mbox{.}}{2015}]%
        {chollet2015keras}
\bibfield{author}{\bibinfo{person}{Fran\c{c}ois Chollet} {et~al\mbox{.}}}
  \bibinfo{year}{2015}\natexlab{}.
\newblock \bibinfo{title}{Keras}.
\newblock \bibinfo{howpublished}{\url{https://github.com/keras-team/keras}}.
  (\bibinfo{year}{2015}).
\newblock


\bibitem[\protect\citeauthoryear{Goodfellow, Bengio, and Courville}{Goodfellow
  et~al\mbox{.}}{2016}]%
        {Goodfellow-et-al-2016}
\bibfield{author}{\bibinfo{person}{Ian Goodfellow}, \bibinfo{person}{Yoshua
  Bengio}, {and} \bibinfo{person}{Aaron Courville}.}
  \bibinfo{year}{2016}\natexlab{}.
\newblock \bibinfo{booktitle}{{\em Deep Learning}}.
\newblock \bibinfo{publisher}{MIT Press}.
\newblock
\newblock
\shownote{\url{http://www.deeplearningbook.org}.}


\bibitem[\protect\citeauthoryear{Hochreiter and Schmidhuber}{Hochreiter and
  Schmidhuber}{1997}]%
        {hochreiter1997long}
\bibfield{author}{\bibinfo{person}{Sepp Hochreiter} {and}
  \bibinfo{person}{J{\"u}rgen Schmidhuber}.} \bibinfo{year}{1997}\natexlab{}.
\newblock \showarticletitle{Long short-term memory}.
\newblock \bibinfo{journal}{{\em Neural computation\/}} \bibinfo{volume}{9},
  \bibinfo{number}{8} (\bibinfo{year}{1997}), \bibinfo{pages}{1735--1780}.
\newblock


\bibitem[\protect\citeauthoryear{Hunter}{Hunter}{2007}]%
        {Hunter:2007}
\bibfield{author}{\bibinfo{person}{J.~D. Hunter}.}
  \bibinfo{year}{2007}\natexlab{}.
\newblock \showarticletitle{Matplotlib: A 2D graphics environment}.
\newblock \bibinfo{journal}{{\em Computing In Science \& Engineering\/}}
  \bibinfo{volume}{9}, \bibinfo{number}{3} (\bibinfo{year}{2007}),
  \bibinfo{pages}{90--95}.
\newblock
\showDOI{%
\url{https://doi.org/10.1109/MCSE.2007.55}}


\bibitem[\protect\citeauthoryear{Loper and Bird}{Loper and Bird}{2002}]%
        {Loper02nltk:the}
\bibfield{author}{\bibinfo{person}{Edward Loper} {and} \bibinfo{person}{Steven
  Bird}.} \bibinfo{year}{2002}\natexlab{}.
\newblock \showarticletitle{NLTK: The Natural Language Toolkit}. In
  \bibinfo{booktitle}{{\em In Proceedings of the ACL Workshop on Effective
  Tools and Methodologies for Teaching Natural Language Processing and
  Computational Linguistics. Philadelphia: Association for Computational
  Linguistics}}.
\newblock


\bibitem[\protect\citeauthoryear{McKinney}{McKinney}{2010}]%
        {mckinney-proc-scipy-2010}
\bibfield{author}{\bibinfo{person}{Wes McKinney}.}
  \bibinfo{year}{2010}\natexlab{}.
\newblock \showarticletitle{Data Structures for Statistical Computing in
  Python}. In \bibinfo{booktitle}{{\em Proceedings of the 9th Python in Science
  Conference}}, \bibfield{editor}{\bibinfo{person}{St\'efan van~der Walt} {and}
  \bibinfo{person}{Jarrod Millman}} (Eds.). \bibinfo{pages}{51 -- 56}.
\newblock


\bibitem[\protect\citeauthoryear{Olah}{Olah}{2015}]%
        {olah2015neural}
\bibfield{author}{\bibinfo{person}{Christopher Olah}.}
  \bibinfo{year}{2015}\natexlab{}.
\newblock \bibinfo{title}{Neural Networks, Types, and Functional Programming}.
\newblock   (\bibinfo{year}{2015}).
\newblock
\showURL{%
\url{http://colah.github.io/posts/2015-09-NN-Types-FP/}}


\bibitem[\protect\citeauthoryear{Pennington, Socher, and Manning}{Pennington
  et~al\mbox{.}}{2014}]%
        {pennington2014glove}
\bibfield{author}{\bibinfo{person}{Jeffrey Pennington},
  \bibinfo{person}{Richard Socher}, {and} \bibinfo{person}{Christopher~D.
  Manning}.} \bibinfo{year}{2014}\natexlab{}.
\newblock \showarticletitle{GloVe: Global Vectors for Word Representation}. In
  \bibinfo{booktitle}{{\em Empirical Methods in Natural Language Processing
  (EMNLP)}}. \bibinfo{pages}{1532--1543}.
\newblock
\showURL{%
\url{http://www.aclweb.org/anthology/D14-1162}}


\bibitem[\protect\citeauthoryear{Services}{Services}{2015}]%
        {emc2015data}
\bibfield{author}{\bibinfo{person}{EMC~Education Services}.}
  \bibinfo{year}{2015}\natexlab{}.
\newblock \bibinfo{booktitle}{{\em Data Science \& Big Data Analytics:
  Discovering, Analyzing, Visualizing and Presenting Data}}.
\newblock \bibinfo{publisher}{John Wiley \& Sons, Inc.}
\newblock


\bibitem[\protect\citeauthoryear{Walt, Colbert, and Varoquaux}{Walt
  et~al\mbox{.}}{2011}]%
        {walt2011numpy}
\bibfield{author}{\bibinfo{person}{St{\'e}fan van~der Walt},
  \bibinfo{person}{S~Chris Colbert}, {and} \bibinfo{person}{Gael Varoquaux}.}
  \bibinfo{year}{2011}\natexlab{}.
\newblock \showarticletitle{The NumPy array: a structure for efficient
  numerical computation}.
\newblock \bibinfo{journal}{{\em Computing in Science \& Engineering\/}}
  \bibinfo{volume}{13}, \bibinfo{number}{2} (\bibinfo{year}{2011}),
  \bibinfo{pages}{22--30}.
\newblock


\bibitem[\protect\citeauthoryear{Waskom, Botvinnik, O'Kane, Hobson, Lukauskas,
  Gemperline, Augspurger, Halchenko, Cole, Warmenhoven, de~Ruiter, Pye, Hoyer,
  Vanderplas, Villalba, Kunter, Quintero, Bachant, Martin, Meyer, Miles, Ram,
  Yarkoni, Williams, Evans, Fitzgerald, Brian, Fonnesbeck, Lee, and
  Qalieh}{Waskom et~al\mbox{.}}{2017}]%
        {michael_waskom_2017_883859}
\bibfield{author}{\bibinfo{person}{Michael Waskom}, \bibinfo{person}{Olga
  Botvinnik}, \bibinfo{person}{Drew O'Kane}, \bibinfo{person}{Paul Hobson},
  \bibinfo{person}{Saulius Lukauskas}, \bibinfo{person}{David~C Gemperline},
  \bibinfo{person}{Tom Augspurger}, \bibinfo{person}{Yaroslav Halchenko},
  \bibinfo{person}{John~B. Cole}, \bibinfo{person}{Jordi Warmenhoven},
  \bibinfo{person}{Julian de Ruiter}, \bibinfo{person}{Cameron Pye},
  \bibinfo{person}{Stephan Hoyer}, \bibinfo{person}{Jake Vanderplas},
  \bibinfo{person}{Santi Villalba}, \bibinfo{person}{Gero Kunter},
  \bibinfo{person}{Eric Quintero}, \bibinfo{person}{Pete Bachant},
  \bibinfo{person}{Marcel Martin}, \bibinfo{person}{Kyle Meyer},
  \bibinfo{person}{Alistair Miles}, \bibinfo{person}{Yoav Ram},
  \bibinfo{person}{Tal Yarkoni}, \bibinfo{person}{Mike~Lee Williams},
  \bibinfo{person}{Constantine Evans}, \bibinfo{person}{Clark Fitzgerald},
  \bibinfo{person}{Brian}, \bibinfo{person}{Chris Fonnesbeck},
  \bibinfo{person}{Antony Lee}, {and} \bibinfo{person}{Adel Qalieh}.}
  \bibinfo{year}{2017}\natexlab{}.
\newblock \bibinfo{title}{mwaskom/seaborn: v0.8.1 (September 2017)}.
\newblock   (\bibinfo{date}{Sept.} \bibinfo{year}{2017}).
\newblock
\showDOI{%
\url{https://doi.org/10.5281/zenodo.883859}}


\end{thebibliography}
